\documentclass{article}

\pdfoutput=1

\usepackage[T1]{fontenc}
\usepackage{microtype}
\usepackage{graphicx}
\usepackage{subfig}
\usepackage{booktabs} 
\usepackage{mathtools}
\usepackage[hidelinks]{hyperref}

\usepackage{graphicx} 

\usepackage[ruled,vlined,algo2e]{algorithm2e}
\usepackage{bbm}
\usepackage{authblk}
\usepackage{fullpage}
\usepackage[round]{natbib}

\usepackage{multirow}
\usepackage{definitions}

\usepackage{xcolor}
\definecolor{mycommentcolor}{HTML}{4f9739}

\SetCommentSty{mycommfont}

\newcommand\numberthis{\addtocounter{equation}{1}\tag{\theequation}}

\newcommand{\eps}{\varepsilon}

\newcommand{\UnnumberedFootnote}[1]{{\def\thefootnote{}\footnote{#1}
\addtocounter{footnote}{-1}}}

\title{Narrowing Action Choices with AI\\Improves Human Sequential Decisions}
\author{Eleni Straitouri$^{*,1}$}
\author{Stratis Tsirtsis$^{*,\S,2}$}
\author{Ander Artola Velasco$^{1}$}
\author{Manuel~Gomez-Rodriguez$^{1}$}

\affil{$^{1}$Max Planck Institute for Software Systems, Kaiserslautern, Germany \\
\{estraitouri, avelasco, manuel\}@mpi-sws.org}
\affil{$^{2}$Hasso Plattner Institute, Potsdam, Germany \\ stratis.tsirtsis@hpi.de}

\date{}

\begin{document}

\maketitle

\UnnumberedFootnote{$^{*}$Authors contributed equally and are listed in alphabetical order.}
\UnnumberedFootnote{$^{\S}$The author contributed to this work during his doctoral studies at the Max Planck Institute for Software Systems.}

\begin{abstract}
Recent work has shown that, in classification tasks, it is possible to design decision support systems that do not require human experts to understand when to cede agency to a classifier or when to exercise their own agency to achieve complementarity---experts using these systems make more accurate predictions than those made by the experts or the classifier alone.
The key principle underpinning these systems reduces to adaptively controlling the level of human agency, by design.
Can we use the same principle to achieve complementarity in sequential decision making tasks?
In this paper, we answer this question affirmatively.
We develop a decision support system that uses a pre-trained AI agent to narrow down the set of actions a human can take to a subset,
and then
asks the human to take an action from this action set.
Along the way, we also introduce a bandit algorithm that leverages the smoothness properties of the action sets provided by our system to efficiently optimize the level of human agency.
To evaluate our decision support system, we conduct a large-scale human subject study ($n = 1{,}600$) where participants play a wildfire mitigation game. 
%
We find that
participants who play the game supported by our system outperform those who play on their own by $\sim$$30$\% and
the AI agent used by our system by $>$$2$\%, even though the AI agent 
largely outperforms participants playing without support.
We have made available the data gathered in our human subject study as well as an open source implementation of our system at \url{https://github.com/Networks-Learning/narrowing-action-choices}.
\end{abstract}

\section{Introduction}
\label{sec:intro}
The idea that humans and machines can work together to achieve greater outcomes than what they can achieve on their own---in short, complementarity---has long intrigued the research community~\citep{jordan1963allocation}. 
Long before the development of modern machine learning, several lines of work provided early evidence that complementarity can be more than just an intriguing idea and may, in fact, be achievable in practice~\citep{brodley1999content, horvitz2007complementary}.   
Since then, a flurry of work has been growing this evidence, especially in the context of automated decision support systems for one-shot prediction tasks~\citep{kamar2012combining, raghu2019algorithmic, choudhury2020machine, de2020regression, bengs2020preselection, de2021classification, mozannar2020consistent, okati2021differentiable, kerrigan2021combining, bansal2021most, steyvers2022bayesian, arnaiz2025towards}.


The conventional wisdom is that, to achieve complementarity in such tasks, human decision makers need to understand when to cede agency to a predictor (\eg, a classifier) or when to exercise their own agency~\citep{yin2019understanding,zhang2020effect,suresh2020misplaced,lai2021towards,bansal2021does, corvelo2023human, corvelo2025human}.
However, a very recent line of work has introduced an alternative family of decision support systems for classification tasks that, perhaps surprisingly, achieve complementarity without such a (strong) requirement~\citep{straitouri2023improving, straitouri2024designing, toni2024towards}.
The key principle underpinning these systems reduces to adaptively controlling the level of human agency---rather than providing a single prediction and letting a human decide when and how to use the prediction, these systems provide a set of predictions, and ask the human to pick one prediction from the set.
%
%
In this work, we investigate whether the same principle can be applied to achieve complementarity in sequential decision making tasks.

%


\xhdr{Our contributions} 
We develop a decision support
system for sequential decision making tasks that uses a pre-trained AI agent to narrow down the set of actions a human decision maker can take to a subset, namely an \emph{action set}, and asks the human to take an action from the action set. 
To develop this system, we start by characterizing how humans make sequential decisions with action sets 
as a structural causal model (SCM)~\citep{pearl2009causality} where, over time, a human collects rewards based on their chosen actions and their resulting effects on the environment's state.
Building upon this characterization,
we show that our decision support system constructs action sets in a way such that the average total reward achieved by a human using our system is Lipschitz continuous with respect to a parameter controlling the size of the action sets and, consequently, the level of human agency.
%
%
Then, we leverage this property to develop an efficient Lipschitz bandit algorithm with sublinear regret guarantees that identifies the optimal level of agency in terms of average total reward, which may be of independent interest as discussed in the comparison with further related work.

Finally, we evaluate our decision support system by conducting a large-scale human subject study, where $1{,}600$ participants play $16{,}000$ instances of a wildfire mitigation game. 
%
%
We find that
participants who play the game supported by our system outperform those who play on their own by $\sim$$30$\% and the AI agent used by our system by $>$$2$\%, even though the AI agent used by our system largely outperforms participants playing without support.

\xhdr{Further related work} 
Our work builds upon further related work on decision support in sequential decision making, multi-armed bandits, and
action masking in reinforcement learning. 

Within the area of decision support in sequential decision making, the work most closely related to ours provides humans with advice while taking actions.
The advice can be in the form of ``tips'' generated by reinforcement learning~\citep{bastani2025improving} or in the form of action recommendations~\citep{grand2024best}.
However, in contrast to our work, humans have the option to ignore the advice and exercise their own agency.
Within this area of research, our work also relates to the notion of algorithmic triage~\citep{straitouri2021reinforcement, zhang2021traded, balazadeh2022learning}, where a human has to take actions for a fraction of the environment states, and an AI agent takes actions for the rest.
%
%

Within the area of multi-armed bandits, our work closely relates to the literature of Lipschitz bandits~\citep{agrawal1995continuum, kleinberg2008multi, bubeck2008online}---refer to~\cite{slivkins2019introduction} for a detailed review on the large family of multi-armed bandit algorithms.
In our work, we view the parameter controlling the level of human agency as an arm of a continuum-armed bandit problem, and show that the average total reward is a Lipschitz function of the arm.
Then, we leverage this property to develop a Lipschitz bandit algorithm for best arm identification with simple regret guarantees. 
Our algorithm can be of independent interest, as it is the first 
to derive simple regret guarantees with no assumptions on the error of the finite sample estimate of the expected payoff of each arm.
%
Existing work on Lipschitz best arm identification with simple regret guarantees~\citep{feng2023lipschitz} makes rather explicit assumptions on the convergence rate of this error. 

Action masking is a technique used by reinforcement learning algorithms that filters---masks out---invalid actions based on the state of the environment (\eg, unavailable actions given the rules of a game) before sampling an action~\citep{vinyals2017starcraft,berner2019dota, ye2020mastering, huang22closer, qian2022dam, muller2022safe, wang2024learning}. 
Action masks can be manually designed based on domain knowledge or learned explicitly using parameterized models, and their goal is to efficiently guide exploration in reinforcement learning algorithms by excluding actions that are irrelevant, unavailable or unsafe given a specific environment.
However, action masks are fundamentally different to our action sets
since they focus on increasing the total reward achieved by algorithms rather than humans.

\section{Sequential Decision Making with Action Sets}
\label{sec:action-sets}
\begin{figure*}
    \centering
    \includegraphics[width=0.6\linewidth]{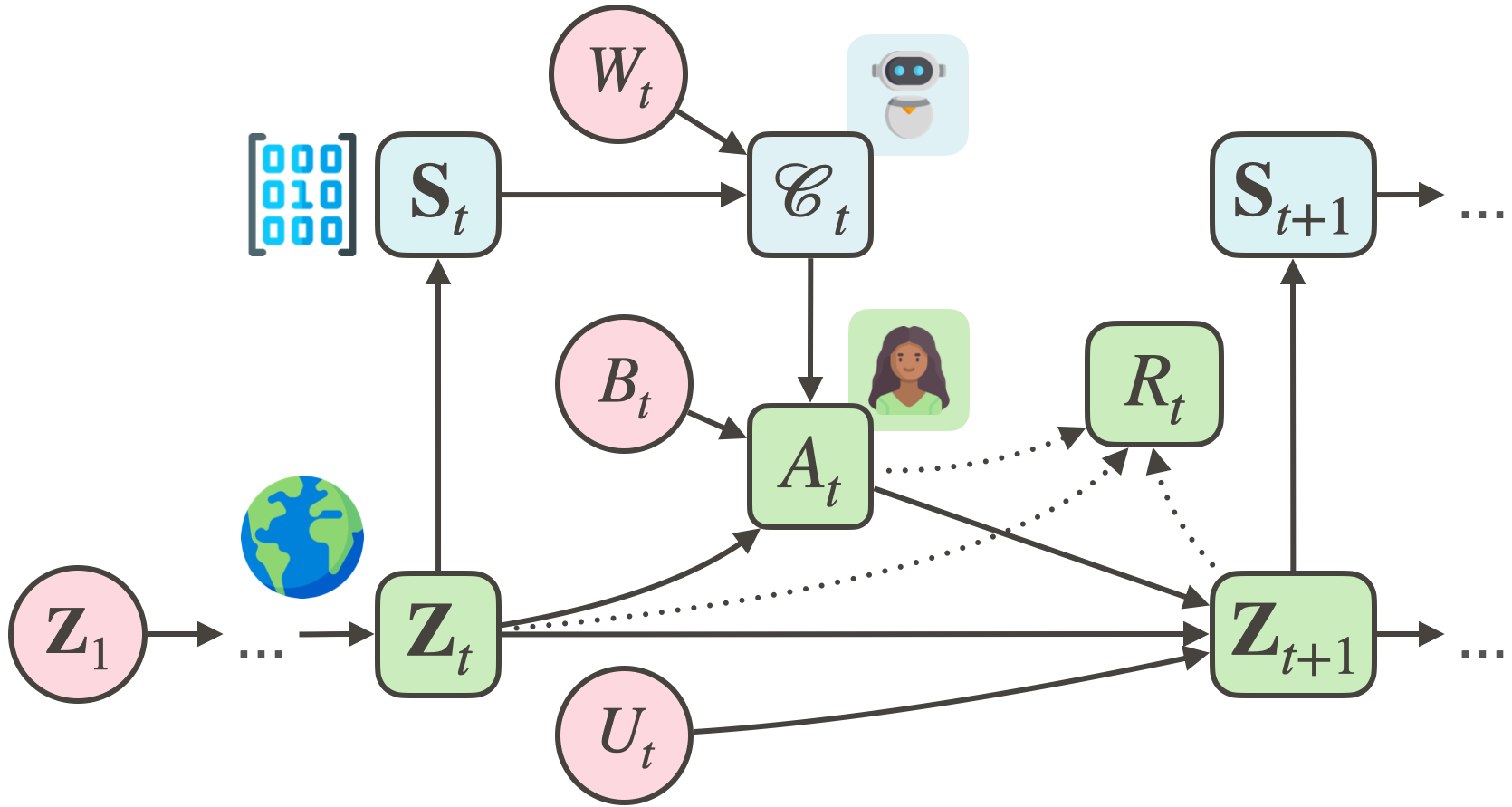}
    \caption{
    \textbf{A causal model of human sequential decision making with action sets.}
    Squares represent endogenous random variables and circles represent exogenous noise variables.
    An arrow from (to) one variable indicates that it is the cause (effect) of another.
    The values of all exogenous variables are sampled from fixed (unknown) distributions,
    while the value of each endogenous variable is given by a function of the values of its ancestors in the causal graph.
    }
    \label{fig:dag}
\end{figure*}

In this section, we first characterize how humans make sequential decisions using our decision support system based on action sets via a structural causal model (SCM)~\citep{pearl2009causality}, building upon previous work on sequential decision making~\citep{oberst2019counterfactual,tsirtsis2021counterfactual} and decision support systems based on prediction sets~\citep{straitouri2024designing}.\footnote{We use upper case letters for random variables, lower case letters for their realizations, calligraphic letters for sets, and bold letters for vectors, unless stated otherwise.}
%
Then, we describe how our decision support system specifically constructs action sets using state-of-the-art AI agents.  

At each time step $t\in\NN$, a human decision maker observes the state of the environment $\Zb_t\in\Zcal$, and our decision support system 
observes a (potentially lossy) representation $\Sbb_t = f_S(\Zb_t)\in\Scal$. 
This captures a variety of real-world scenarios in which the human decision maker has access to salient but hard-to-quantify information that makes complementarity a particularly worthy goal~\citep{alur2023auditing}; for example, a doctor's direct conversations with patients in healthcare, a scout's qualitative assessment of prospective players in professional sports, or an officer's handling of resource constraints in disaster response.
%
Then, given a finite space of $m$ possible actions $\Acal$, our decision support system
narrows down the space of actions the human can take to an action set 
\begin{equation}\label{eq:action-sets}
   \Ccal_t = \pi_\Ccal(\Sbb_t, W_t) \subseteq \Acal,
\end{equation}
where $\pi_\Ccal$ is a decision support policy, and $W_t \sim P(W)$ is an (exogenous) noise variable that allows for randomized action sets.\footnote{In Section~\ref{sec:lipschitz}, we leverage this randomization to optimize the decision support policy $\pi_\Ccal$ using Lipschitz bandit algorithms.}
Further, given the action set $\Ccal_t$ provided by our decision support system 
and the current state $\Zb_t$ of the environment, the human takes an action
\begin{equation}\label{eq:human-policy}
   A_t = \pi_H(\Zb_t, \Ccal_t, B_t),
\end{equation}
where $\pi_H$ is an (unknown) human policy, and $B_t$ is a noise variable that captures randomness in human decisions.
%
%
Subsequently, depending on the action $A_t$ taken by the human and the current state $\Zb_t$, the environment transitions to the next state 
%
$\Zb_{t+1} = f_Z(\Zb_t, A_t, U_t)$,
%
where $U_t$ is a noise variable that captures inherent randomness in the environment's state transitions.\footnote{
Note that the state $\Zb_t$ may include the entire history of past states and actions, and the noise variables $\Ub_t$ may be correlated across time steps. As a consequence, our model also allows for non-Markovian environment dynamics and human policies.}
Finally, following each transition, the human receives a numerical reward $R_t = f_R(\Zb_t, A_t, \Zb_{t+1}) \in [-r_\text{max},r_\text{max}]$.
%
%
Figure~\ref{fig:dag} shows a visual representation of our causal model.

To define the decision support policy $\pi_\Ccal$ and in turn construct action sets $\Ccal_t$, 
our decision support system utilizes an AI agent'{}s valuation $q(\sbb, a)$ of each action $a \in \Acal$ given a representation $\sbb \in \Scal$.
The AI agent'{}s valuation may correspond, for example, to Q-values given by a deep Q-network in the context of reinforcement learning~\citep{mnih2013playing} or logits over a set of possible actions given by a large language model fine-tuned for sequential decision making~\cite{li2022pre}.\footnote{Whenever AI agents operate autonomously, they often implement a greedy policy $\pi_M(\sbb) = \argmax_{a\in\Acal} q(\sbb, a)$. However, if $\sbb$ is a lossy representation of $\zb$, such a greedy policy is likely to be suboptimal.}
More specifically, let $a_{(1)}, a_{(2)}, ..., a_{(m)}$ be the ranking of the actions $a\in\Acal$ according to the AI agent'{}s valuation $q(\sbb, a)$. 
%
Then, the decision support policy $\pi_{\Ccal}$ is given by
\begin{align}\label{eq:decision-support-policy}
 \pi_\Ccal(\sbb, W ;\eps) = \{a_{(i)}\}_{i=1}^{k}, \,\text{where }    k = 1 +  \sum_{j=2}^{j = m} 
 \mathbbm{1}\left\{
 \tilde{q}(\sbb, a_{(j)}) + W \geq \tilde{q}(\sbb, a_{(1)}) - \eps
 \right\}, 
\end{align}    
%
%
and $W$ is a half-normal noise variable $W = |X|$ with $X\sim \Ncal(0,\sigma^2)$, 
$\tilde{q}(\sbb, a) \in [0,1]$ denotes a scaled version of the original valuation $q(\sbb, a)$ computed via min-max normalization,
%
and $\eps \in [0, 1]$ is a parameter controlling the size of the action set and, consequently, the level of human agency.

Intuitively, the above decision support policy gives a small random boost $W$ to the (scaled) scores of all actions except for the best one, and it includes in the action set all actions whose boosted scores are within $\eps$ of the score of the best action.
Our motivation for this construction is twofold.
First, it allows us to control the level of human agency using a (continuous) parameter that is independent of the size of the action space $\Acal$.
For example, for $\eps = 1$, the action set contains all actions $\Acal$ and thus the human maintains full agency and, for $\eps = 0$ and $\sigma$ sufficiently small, the action set contains only the highest ranked action $a_{(1)}$ and the human has no agency.
Second, the randomization introduced by the random variable $W$ ensures that, as we will show later on, the probability distribution over candidate action sets entailed by Eq.~\ref{eq:decision-support-policy} varies smoothly with $\eps$.
In what follows, we leverage this smoothness property to develop a bandit algorithm that efficiently finds the optimal value of the parameter $\eps$ controlling the level of human agency that maximizes the (discounted) cumulative reward, \ie,
\begin{equation}\label{eq:cumulative-reward}
    \eps^{*} = \argmax_{\eps \in [0,1]}
    \EE_{\Zb_1, \Wb, \Bb, \Ub} \left [ \sum_{t=0}^{+\infty} \gamma^{t} \cdot R_{t+1}  \right],
\end{equation}
where $\gamma \in (0,1)$ is a given discount factor, and the expectation is over the initial state of the decision making process $Z_1$ and the randomness in the decision support policy, the human'{}s policy, and the environment's state transitions.
%

\section{A Lipschitz Bandit Algorithm to Identify Optimal Action Sets}
\label{sec:lipschitz}
\begin{algorithm}[t]
\caption{Lipschitz Best Arm Identification} 
\label{algo:best-arm-identification}
\SetAlgoLined
\textbf{Input:} Exploration budget $n$, Lipschitz constant $L$, exploitation parameter $\beta$.\\
$t_{1} \leftarrow 0$, $k \gets 1$, $\Ical_{1} \leftarrow \left \{[0,1/2], [1/2, 1] \right \}$\;

\While{$t_{k} \leq n$}{
    $l_{k} \leftarrow 2^{-k}$\;
    
    $n_k \leftarrow 2^{k  \beta}$\;
    
    \For{$I \in \Ical_k$}{
        $\eps_{I} \gets \mathrm{midpoint}(I)$ \;
        
        $\hat{p}(\eps_I) \leftarrow \textsc{PullArm}(\eps_I, n_{k})$ $\,\,\,$\tcp{Pull the midpoint of each interval $n_k$ times}
    }  
    
    $\hat{p}_{\text{max}} \gets \max_{I \in \Ical_{k}} \hat{p}(\eps_I)$\;
    
    $\eps_{\text{OPT}} \gets \textsc{Midpoint}\left(\argmax_{I \in \Ical_{k}} \hat{p}(\eps_I)\right)$\\ $\,\,\,$\tcp{Find the interval whose midpoint achieves the highest average payoff}
    
    $\Ical_{k+1} \gets \varnothing$ \;
    
    \For{ $I \in \Ical_{k}$}{
        \tcp{Zoom-in and explore further only intervals whose average payoff is close to the highest}
        \If{$\hat{p}_{\text{max}}-\hat{p}(\eps_I)\leq (2+L/2)l_k $ }
        {
            $I_{\text{low}}, I_{\text{high}} \gets \textsc{PartitionInHalf}(I)$ \;
            
            $\Ical_{k+1} \gets \Ical_{k+1} \cup \{ I_{\text{low}}, I_{\text{high}}\}$ \;
        } 
    }
    
    $t_{k+1}\gets t_{k} + n_{k}|\Ical_{k}|$\;
    
    $k \gets  k + 1$\;
}
\textbf{return} $\eps_{\text{OPT}}$
\end{algorithm}
\begin{figure}[t]
    \centering
    \includegraphics[width=0.5\linewidth]{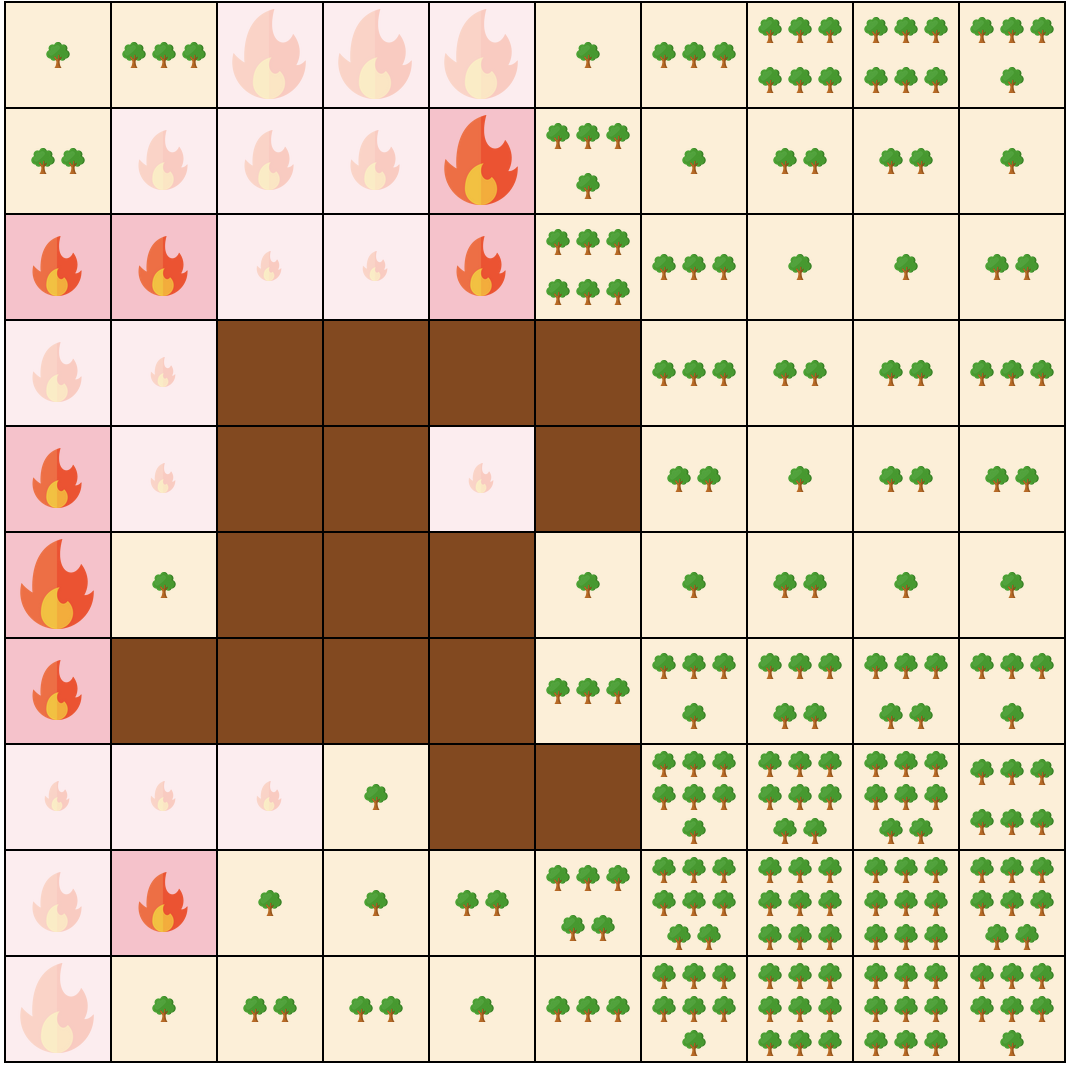}
    \caption{\textbf{Instance of our wildfire mitigation game as shown to human participants.} Tiles with green trees are \texttt{healthy}, tiles with fire icons are \texttt{burning}, and brown tiles are \texttt{burnt}.
    The size of the fire icon in each \texttt{burning} tile is proportional to the time steps left until the tile turns to \texttt{burnt} on its own.
    Non-faded \texttt{burning} tiles correspond to the action set given by our decision support system, while faded tiles correspond to actions that the decision support system does not allow the human participant to select.
    }
    \label{fig:game-example}
\end{figure}
Our starting point is the observation that, based on Eq.~\ref{eq:decision-support-policy}, a decision support policy can assign non-zero probability only to $m$ different action sets---those of the form $\Ccal_{(i)} = \{ a_{(1)}, a_{(2)}, \ldots, a_{(i)} \}$, where $a_{(1)}, a_{(2)}, \ldots, a_{(m)}$ is the ranking of the $m$ actions based on the valuations $q(\sbb, a)$ of the AI agent. 

Building upon that observation, we first show that the total variation distance between two distributions over the $m$ action sets induced by two different values of the parameter $\eps$ is bounded (up to a multiplicative factor) by the absolute difference between those two values.\footnote{The proofs for all propositions can be found in Appendix~\ref{app:proofs}.}
\begin{proposition}~\label{prop:half-normal}
    Let $L_c = \frac{2\sqrt{2}}{\sigma \sqrt{\pi}}$. For any representation $\sbb \in \Scal$ and any $\eps, \eps' \in [0,1]$, it holds that
    \begin{align}
        \sum_{i=1}^{m}
        \left|  \PP[\pi_{\Ccal}(\sbb, W; \eps) = \Ccal_{(i)} ]  \right . \left .  -  \PP[\pi_{\Ccal}(\sbb, W; \eps') = \Ccal_{(i)} ] \right|  \leq L_c |\eps - \eps'|.
    \end{align}
\end{proposition}

Perhaps surprisingly, 
the above proposition suggests that the amount by which the two distributions can differ depends only on the standard deviation of the half-normal random variable $W$ used in our decision support policy and not on the number of actions $m$.
Further, we leverage the proposition 
to show that the (discounted) cumulative reward achieved by a human using our decision support policy varies smoothly with respect to the parameter $\epsilon$.
More formally, let 
\begin{equation*}
v(\zb ; \eps) = \EE_{\Wb, \Bb, \Ub} \left [ \sum_{t=0}^{+\infty} \gamma^t \cdot R_{t+1} \given \Zb_{1} = \zb\right]
\end{equation*}
denote the (discounted) cumulative reward achieved by a human using a decision support policy $\pi_{\Ccal}$ parameterized by $\eps$, conditioned on an initial state $\zb$. 
Then, the following proposition
suggests that its expectation over the initial state $\Zb_{1}$ is Lipschitz-continuous with respect to $\eps$.
%
\begin{proposition}~\label{prop:lipischitz-v}
    Let $L_v =  \frac{L_{c}\cdot r_{max}}{(1 - \gamma)^2}$. For any $\eps, \eps' \in [0,1]$, it holds that 
    \begin{equation}
        \left | \EE_{\Zb_{1}}[v(\Zb_{1};\eps)] - \EE_{\Zb_{1}}[v(\Zb_{1};\eps')] \right|
        \leq L_v \cdot | \eps - \eps' |. 
    \end{equation}
\end{proposition}

As an immediate consequence, we can find the optimal value of the parameter $\eps^*$ defined in Eq.~\ref{eq:cumulative-reward} by viewing each parameter value $\eps\in[0,1]$ as an arm in a best-arm identification Lipschitz bandit problem~\citep{agrawal1995continuum, kleinberg2008multi, bubeck2008online}.
In a typical best-arm identification problem, one pulls different arms repeatedly and observes a (sampled) payoff for each pull. Based on these observations, the objective is to identify the arm that yields the maximum expected payoff.
In our setting, we view the act of pulling an arm $\eps$ as a single instance of our sequential decision making task, in which the human takes a sequence of actions 
constrained by the 
action sets generated by our decision support policy with the parameter $\eps$.
Further, the observed payoff for each pull is the cumulative (discounted) reward achieved by the human in the respective instance, leading to an expected payoff for an arm $\eps$ equal to $p(\eps) = \EE_{\Zb_1} \left[v(\Zb_1;\eps)\right]$.\footnote{In practice, to be able to observe the payoff for each pull, the environment transitions need to be such that the decision making process eventually terminates, that is, it reaches an absorbing state that yields zero reward.}
%
In what follows, we develop a best-arm identification algorithm that pulls arms efficiently by leveraging the Lipschitz-continuity of the expected payoffs shown in Proposition~\ref{prop:lipischitz-v}.

Our algorithm, summarized in Algorithm~\ref{algo:best-arm-identification}, proceeds in iterations and it uses a technique known as \emph{zooming}~\citep{kleinberg2008multi}.
Specifically, in each iteration $k$, it starts with a set $\Ical_k$ of \textit{active} (sub-)intervals of $[0,1]$, that is, intervals that are more likely to include the best arm $\eps^*$.
Then, it 
explores each active interval $I \in \Ical_{k}$
by pulling $n_{k}$ times an arm $\eps_{I}$ representative of the interval $I$ (\eg, its midpoint) and keeping track of its empirical average payoff $\hat{p}(\eps_{I})$.
Based on a given Lipschitz constant $L$ for the expected payoffs, the algorithm continues by identifying and eliminating intervals that are unlikely to include the best arm.
Lastly, it partitions the remaining intervals in half to construct the new set of active intervals to be explored in the next iteration.
The algorithm terminates once it consumes a user-specified exploration budget of $n$ arm pulls. 

The aforementioned procedure allows us to derive guarantees on the performance of our algorithm in terms of its \emph{simple regret}, that is, the difference between the expected payoff of the optimal arm $\eps^*$ and the arm $\eps_{ALG}$ returned by our algorithm.
Importantly, the guarantees we derive depend on an instance-specific number $d\in\NN$, known as the \emph{zooming dimension}~\citep{kleinberg2008multi}, which captures how ``difficult'' it is to identify the best arm in a given Lipschitz bandit problem instance---low values of $d$ indicate that arms with close-to-optimal expected payoffs are concentrated around the best arm, while high values of $d$ indicate that such arms are spread out in the $[0,1]$ interval.\footnote{Formally, the zooming dimension of a bandit problem is the smallest  $d>0$ such that there exists a constant $\lambda>0$ with the following property: for every $\rho>0$, there exist $N \leq \lambda \rho^{-d}$ intervals $I \subseteq [0,1]$ of length $\rho$ that cover the set of arms $\{\eps\in[0,1]: p(\eps^*) - p(\eps) \leq \rho\}$.}
Formally, we have the following proposition regarding the simple regret of Algorithm~\ref{algo:best-arm-identification}:
\begin{proposition}\label{prop:simple-regret}
    Let $d$ be the zooming dimension of an instance of the problem given by Eq.~\ref{eq:cumulative-reward}. For any $n \geq 1$, there exists a $\delta_n>0$ such that, with probability at least $1 - \delta_n$,
    Algorithm \ref{algo:best-arm-identification} with exploitation parameter $\beta$ returns an arm $\eps_{ALG}$ that satisfies
    \begin{equation*}
        p(\eps^*) - p(\eps_{ALG})
        \leq \Ocal(n^{\frac{-1}{d+\beta}}).
    \end{equation*}
\end{proposition}

The inequality above tells us that the simple regret achieved by our algorithm improves as the exploration budget $n$ increases.
Moreover, the rate at which this happens depends on the zooming dimension $d$, with lower values of $d$ (\ie, easier instances) leading to faster rates of decrease in the simple regret.
Although we have omitted the constants in the expression above for brevity, note that the simple regret also improves with lower values of the given Lipschitz constant $L$ (refer to Eq.~\ref{eq:full-regret} in Appendix~\ref{app:simple-regret} for the full expression).
In what follows, we evaluate the performance of Algorithm~\ref{algo:best-arm-identification} in efficiently identifying the optimal value of the parameter $\eps$ in a human subject study, along with the overall ability of our decision support policy introduced in Section~\ref{sec:action-sets} to improve human sequential decisions.

\section{Evaluation via Human Subject Study}
\label{sec:experiments}
\begin{figure}[t]
    \centering
    \includegraphics[width=.5\linewidth]{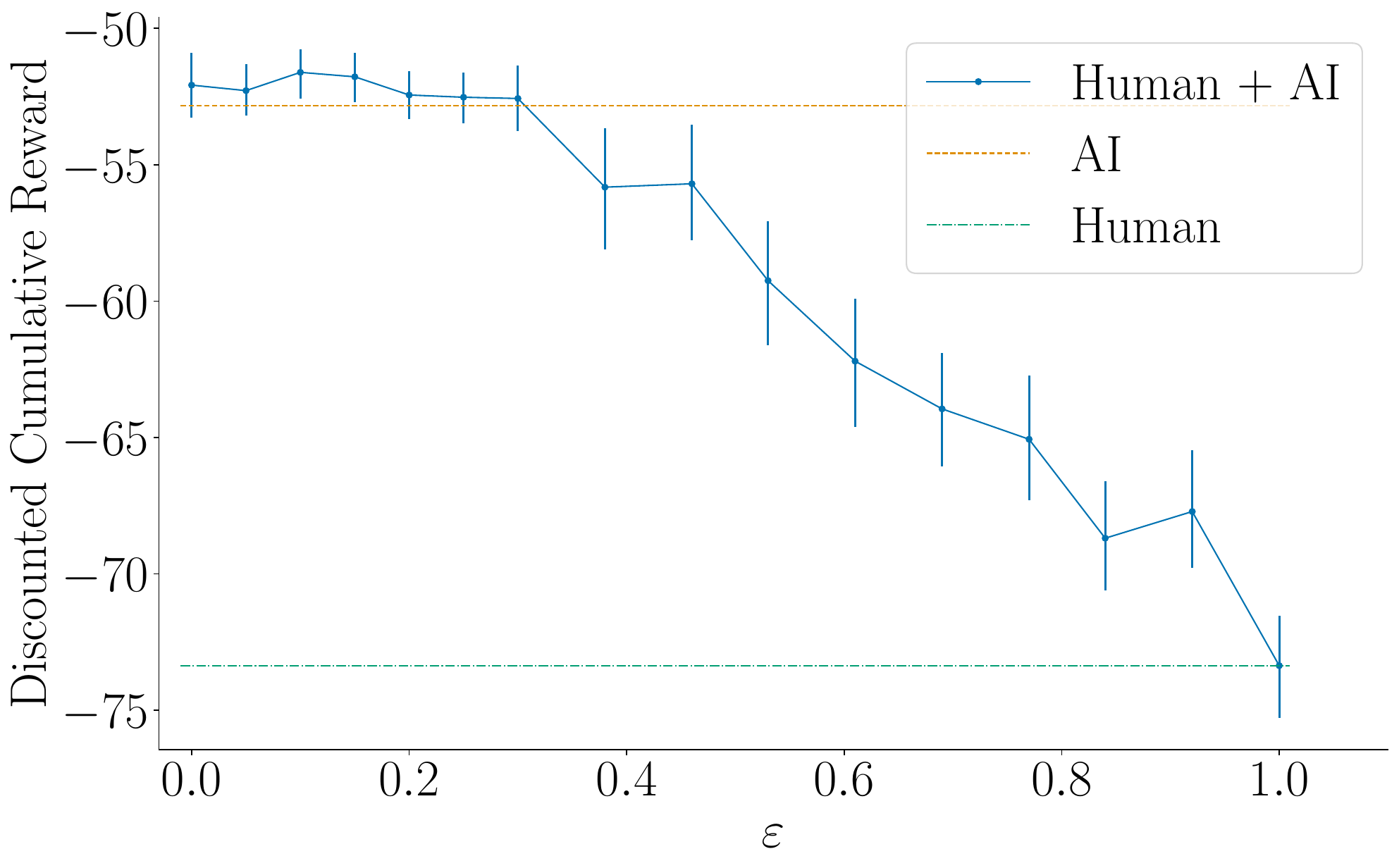}
    \caption{\textbf{Discounted cumulative reward achieved by human participants supported by our system, human participants on their own, and the AI agent used by our system.} 
    In the blue line, each point correspond to a different value of the parameter $\eps$ controlling the level of human agency and error bars correspond to $95\%$ confidence intervals.
    Here, we use $\gamma=0.99$ and, to improve visibility, in the range of $\eps\in[0,0.3]$, we create coarse-grained bins of $\eps$ of size $5$ and, for each bin, we show the average discounted cumulative reward.
    }
    \label{fig:human-vs-dqn}
\end{figure}
\begin{figure}[t]
    \centering
    \includegraphics[width=0.47\linewidth]{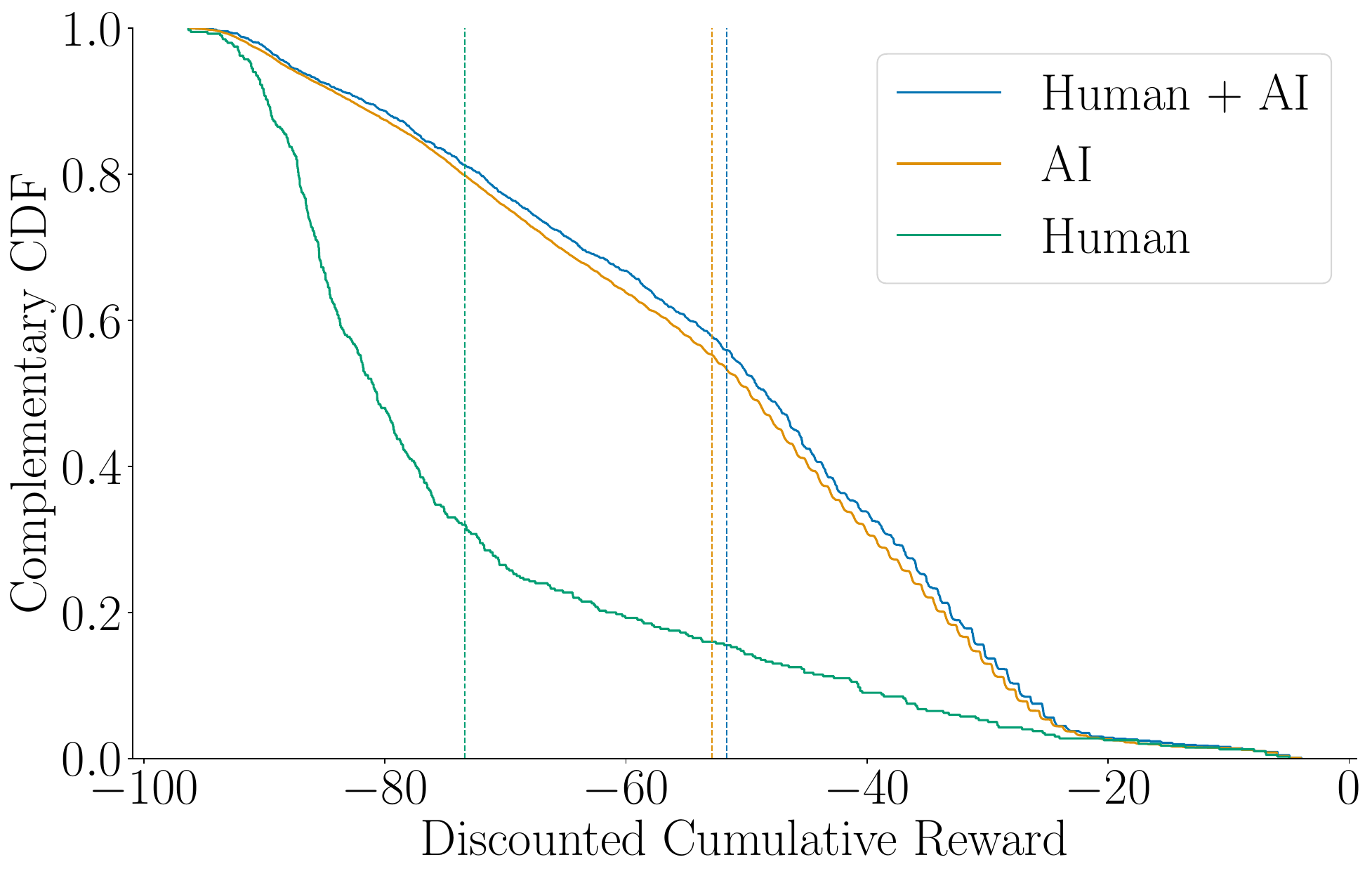}
    \caption{
    \textbf{Distribution of discounted cumulative rewards achieved by human participants supported by our system, human participants on their own, and the AI agent used by our system.}
    %
    %
    Within the plot, vertical lines correspond to the respective empirical average discounted cumulative reward. 
    Here, our decision support system uses $\eps \in [0.08, 0.12]$, and
    a t-test shows that its difference from the respective distribution of the AI agent alone and the average human alone is statistically significant with p-value $0.01$.}
    %
    \label{fig:human-vs-dqn-ccdf}
\end{figure}

In this section, we develop a wildfire mitigation game and use it to conduct a large-scale human subject study.
Our goal is to show that: 
i) under the optimal value of the parameter $\eps$ controlling the level of human agency, our system achieves human-AI complementarity and, 
%
ii) our Lipschitz best arm identification algorithm 
successfully identifies the optimal value of $\eps$, while achieving a notable improvement in simple regret compared with a competitive baseline.\footnote{We ran our experiments on a Debian machine with an ARM EPYC 7662 processor, 20 cores and 32GB memory.}     

\xhdr{Human subject study setup} We recruited $1{,}600$ participants through Prolific\footnote{\href{https://www.prolific.com/}{https://www.prolific.com/}}. 
%
We compensated 
%
each participant with $9$ GBP per hour 
pro-rated, as recommended by Prolific, and 
we doubled the compensation of the top $10\%$ of participants in terms of their score in our game 
to further incentivize good performance.
Each participant provided their consent to participate by filling in a 
form with a detailed description of the purpose and process of the study.  
Our experimental protocol received approval by the Institutional Review Board (IRB) from the University of Saarland. 

Each participant went through a short tutorial and then played $10$ instances of our wildfire mitigation game. 
In each game instance, we collected data at each individual time step that included the state of the game, the action that the participant took, and the immediate (in-game) reward they received.
We did not collect personally identifiable information and stored the data mentioned above in an anonymized way.
For further details regarding our human subject study setup, refer to Appendix~\ref{app:experiment-details}.

\xhdr{Wildfire mitigation game} Each game instance includes a fictional forest map with an on-going wildfire, and the participant's goal is to prevent it from spreading.
Figure~\ref{fig:game-example} shows a snapshot of one of the game instances played by a participant.

The \emph{state} of the game is characterized by the status of the fictional forest, shown as a $10$x$10$ grid, where each grid tile is either \texttt{healthy} (has green trees), 
\texttt{burning} (on fire), or \texttt{burnt} (brown).
The number of trees in each \texttt{healthy} tile indicates how easy it is for the tile to catch fire (\ie, more trees indicate an easier spread).
Moreover, the size of the fire in each \texttt{burning} tile indicates how long the fire in that tile can keep burning (\ie, smaller fires will extinguish themselves sooner).

At each time step of the game, our system 
creates an \emph{action set} that contains a subset of the \texttt{burning} tiles (non-faded) and the participant takes an \emph{action} by selecting 
one of those tiles to which they can apply water mitigation measures;
these extinguish the fire and turn the respective tile 
to \texttt{burnt}. 

The \emph{transitions} between states and the immediate \emph{rewards} are determined by the actions of the participant and the dynamics of the fire spread.
At each time step, each \texttt{burning} tile spreads the fire stochastically to a subset of its neighboring \texttt{healthy} tiles (refer to Appendix~\ref{app:experiment-details} for details), and the immediate reward equals the negative total number of tiles that caught fire in that time step.
In addition, a \texttt{burning} tile to which no water mitigation measures have been applied turns to \texttt{burnt} on its own after three time steps.
Lastly, once a tile becomes \texttt{burnt} it cannot become neither \texttt{burning} nor \texttt{healthy} again.     
The game ends when there are no more  \texttt{burning} tiles left, and the participant gets a score equal to the number of remaining \texttt{healthy} tiles.\footnote{We use $\gamma\approx 1$, hence, the (discounted) cumulative reward is approximately equal to $score - 100$.}

\xhdr{Decision support policies} We implement the decision support policies underpinning our system, defined by Eq.~\ref{eq:decision-support-policy}, using a range of parameter values $\eps$ from $0.0$ to $0.3$ with step $0.01$, and from $0.3$ to $1.0$ with step $0.05$.
For each $\eps$ value, we create a set of $400$ game instances in which participants play the wildfire mitigation game supported by our decision support system, and each participant is randomly assigned $10$ instances with multiple $\eps$ values.
As an AI agent, we train and use a Deep Q-Network (DQN)~\citep{mnih2015human} that, at each time step of a game instance, returns valuations $q(\sbb, a)$ for all actions $a$ (\ie, \texttt{burning} tiles) depending on the current representation $\sbb$ (\ie, the state of the game).
%
%
%
Lastly, for the half-normal distribution of the random variable $W$ that adds stochasticity to the construction of action sets, we use $\sigma=0.01$. 
For further details on the implementation of our decision support policies as well as the training of our Deep Q-Network refer to Appendix~\ref{app:experiment-details}.

\xhdr{Results}
Figure~\ref{fig:human-vs-dqn} provides evidence that our decision support system can achieve complementarity under se\-ve\-ral values of the parameter $\eps$ controlling the level of human agency. 
In particular, under the optimal value of the parameter $\eps$, a human using our decision support system achieves an average (discounted) cumulative reward $29.65\%$ higher than the one a human achieves by playing the game on their own (\ie, under $\eps=1$).
Perhaps surprisingly, even though the AI agent used by our system outperforms human participants playing on their own by a large margin, 
our decision support system under the optimal value of $\eps$ manages to further improve its average (discounted) cumulative reward by $2.31\%$.
Figure~\ref{fig:human-vs-dqn-ccdf} further supports this evidence, by showing that the empirical cumulative distribution of the (discounted) cumulative reward under the optimal value of $\eps$ not only differs from the distributions of the (discounted) cumulative reward achieved by the Deep Q-Network and humans playing on their own, but also dominates them. To further support this claim, we ran a t-test, which shows that the difference between the distributions is statistically significant (p-value $0.01$).
%

\begin{figure}[t]
    \centering
    \includegraphics[width=0.5\linewidth]{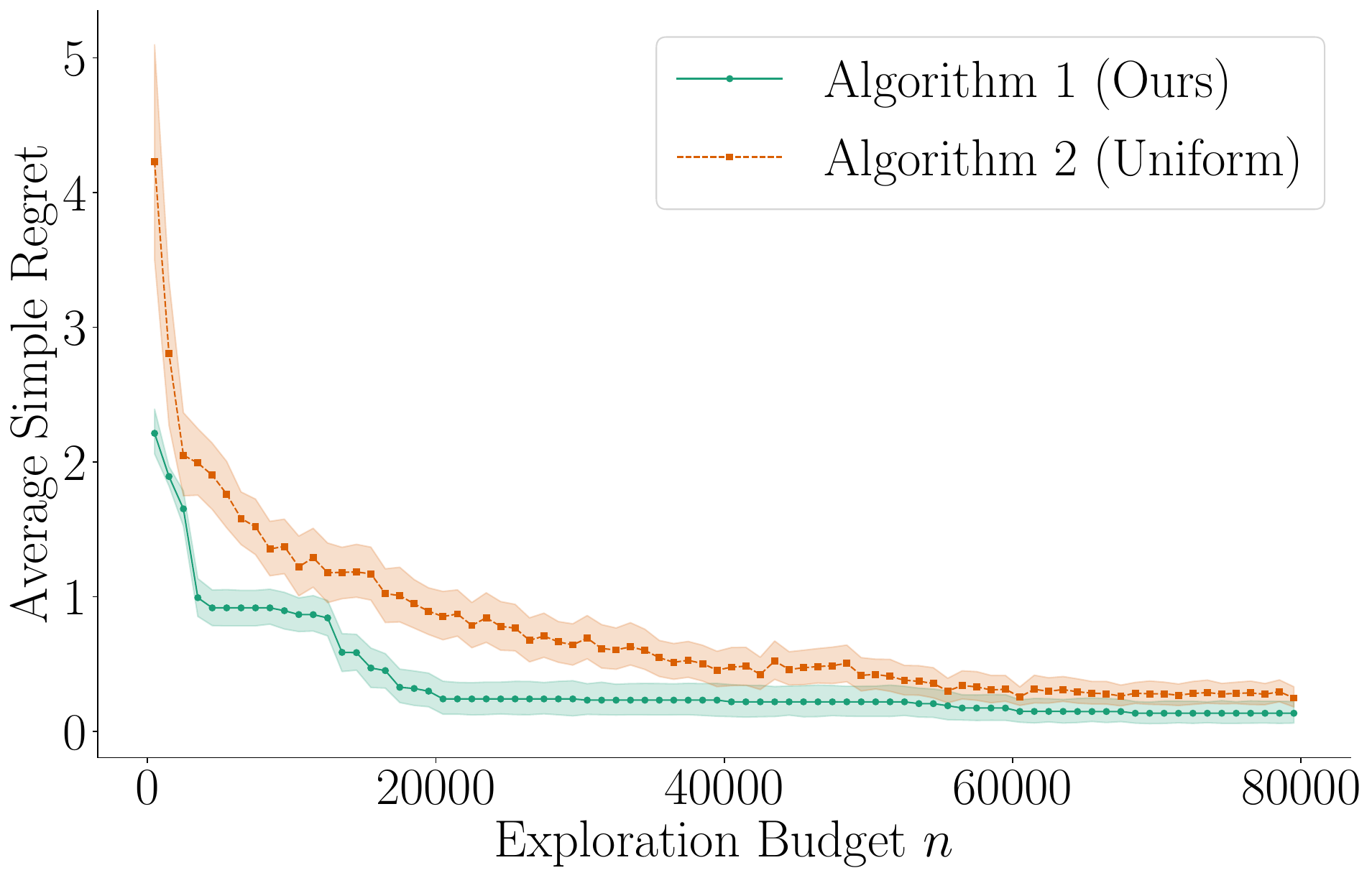}
    \caption{\textbf{Simple regret achieved by our Lipschitz best-arm identification algorithm against a uniform discretization baseline.}
    Each point in the green (orange) line shows the empirical average simple regret achieved over $100$ executions of Algorithm~\ref{algo:best-arm-identification} (\ref{algo:uniform}), respectively, 
    given the respective exploration budget $n$.
    Here, we repeat each experiment $100$ times using a different random seed for the payoffs obtained from arm pulls and shaded areas represent $95\%$ confidence intervals.
    In Algorithm~\ref{algo:best-arm-identification}, we set $\beta=2$ and $L=150$.}\label{fig:regret}
\end{figure}
\begin{figure}[t]
    \centering
    \includegraphics[width=.5\linewidth]{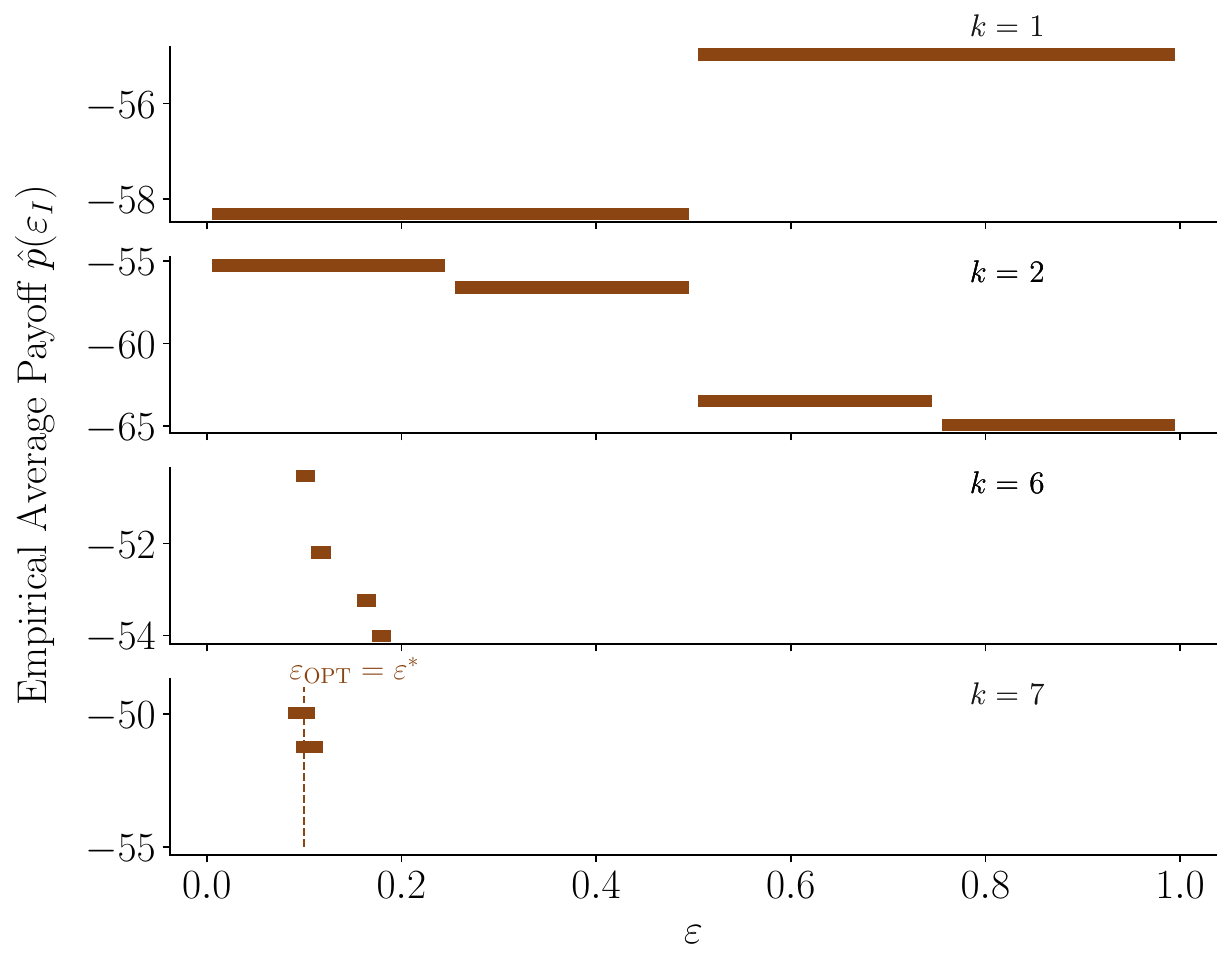}
    \caption{\textbf{Evolution of active intervals throughout the execution of Algorithm~\ref{algo:best-arm-identification}.}
    From top to bottom, the different plots correspond to the first two and last two iterations of one execution of Algorithm~\ref{algo:best-arm-identification}.
    The length of each brown bar on the $x$-axis corresponds to a range of $\eps$ values that form an active interval  and its value on the $y$-axis corresponds to the empirical average payoff the algorithm has obtained by pulling the arm corresponding to that active interval.
    In the bottom plot, the dashed line indicates the (optimal) value  $\eps_{OPT} = \eps^*$, which was successfully identified by the algorithm.
    Here, we set the parameters of Algorithm~\ref{algo:best-arm-identification} to $n=30{,}000$, $\beta=2$, and $L=150$.
    }
    \label{fig:intervals}
\end{figure}

Further, we evaluate the performance of Algorithm~\ref{algo:best-arm-identification} in identifying the optimal arm $\eps^*$ using our human subject study data.
To this end, we compare its simple regret against the simple regret achieved by the uniform discretization algorithm~\citep{slivkins2019introduction} under several values of the exploration budget $n$. For completeness, we include the uniform discretization algorithm under Algorithm~\ref{algo:uniform}.
Here, we implement Algorithm~\ref{algo:uniform} using $100$ discretization levels, \ie, $100$ equally sized intervals covering the entire range of $\eps$ values and distribute the exploration budget uniformly across those. 
Figure~\ref{fig:regret} summarizes the results, which show that the simple regret of our algorithm converges to zero as the exploration budget $n$ increases, as expected.
Moreover, they also show that, for the same exploration budget $n$, our algorithm achieves lower simple regret compared to the uniform discretization algorithm.
Figure~\ref{fig:intervals} complements these results by showing the active intervals in the first and last two iterations for one execution of Algorithm~\ref{algo:best-arm-identification}. 
The results demonstrate that, in practice, Algorithm~\ref{algo:best-arm-identification} successfully zooms in the interval containing the optimal value $\varepsilon^*$, and identifies it as the optimal.

\section{Discussion and Limitations}
\label{sec:discussion}
In this section, we highlight several limitations of our work and discuss avenues for future research as well as its broader impact.

\xhdr{Methodology}
%
%
Our decision support system utilizes the same parameter value $\eps$ for each state of the environment. 
However, one may obtain improved performance by utilizing a different parameter value $\eps(\sbb)$ per state $\sbb$.
%
%
%
By design, our methodology ensures that a human who takes action supported by our decision support system achieves better or equal average performance than the human or the AI agent used by our system on their own. 
However, it would be very important to better understand the factors that influence the degree of human-AI complementarity offered by our system.
%
Finally, our decision support system requires actions to be discrete. 
It would be interesting to lift this requirement and allow for continuous actions.

\begin{algorithm}[!t]
\caption{Uniform Discretization} 
\label{algo:uniform}
\SetAlgoLined
\textbf{Input:} Exploration budget $n$, discretization levels $\Dcal$.\\
    \For{$I \in \Dcal$}{
        $\eps_{I} \gets \mathrm{midpoint}(I)$ \;
        
        $\hat{p}(\eps_I) \leftarrow \textsc{PullArm}(\eps_I, n / |\Dcal|)$ $\,\,\,$\tcp{Pull the midpoint of each interval $n / |\Dcal| $ times}
    }  
    
    $\hat{p}_{\text{max}} \gets \max_{I \in \Dcal} \hat{p}(\eps_I)$\;
    
    $\eps_{\text{OPT}} \gets \textsc{Midpoint}\left(\argmax_{I \in \Dcal} \hat{p}(\eps_I)\right)$\\ $\,\,\,$\tcp{Find the interval whose midpoint achieves the highest average payoff}
\textbf{return} $\eps_{\text{OPT}}$
\end{algorithm}

\xhdr{Evaluation}
%
Our large-scale human subject study suggests our decision support system may enable human-AI complementarity in sequential decision making tasks. 
However, it comprises only one sequential decision making task---a toy wildfire mitigation game---and thus one may question its generalizability. 
It would be important to conduct additional human subject studies comprising other (real) sequential decision making tasks. 
However, it is worth highlighting that conducting such studies at scale would entail significant financial costs---the total cost of our human subject study  was $5{,}280$ GBP.

\xhdr{Broader Impact}
Our work has focused on maximizing the average cumulative reward achieved by a human using a decision support system based on action sets. 
However, whenever the decision support system is used in high-stakes domains, it would be important to extend our methodology to account for safety and fairness considerations.

\section{Conclusions}
\label{sec:conclusions}
In this paper, we have developed a decision support system for sequential decision making tasks that do not require human experts to understand when to cede agency to the system to achieve human-AI complementarity.
Along the way, to optimize the design of our system, we have also introduced and analyzed a novel Lipschitz bandit algorithm with sublinear regret guarantees, which may be of independent interest.
Further, we have conducted a large-scale human subject study with thousands of participants to experimentally validate that our system achieves human-AI complementarity in a (non-trivial) sequential decision making task---a wildfire mitigation game.

\vspace{2mm}
\xhdr{Acknowledgements} 
Gomez-Rodriguez acknowledges support from the European Research Council (ERC) under the European Union'{}s Horizon 2020 research and innovation programme (grant agreement No. 945719). Straitouri acknowledges support from a Google PhD Fellowship. We would also like to give credit to the creator Freepik from
\url{https://www.flaticon.com} whose icons we have used to design our experiment.

{ 
\small
\bibliographystyle{unsrtnat}
\bibliography{action-sets}
}

\clearpage
\newpage

\appendix

\section{Proofs}\label{app:proofs}

\subsection{Proof of Proposition~\ref{prop:half-normal}}


Given a ranking $a_{(1)}, a_{(2)}, \ldots, a_{(m)}$ of the $m$ actions based on the (scaled) valuations $\tilde{q}(\sbb, a)$ of the AI agent, let $\Delta_i$ be the difference between the highest and the $i$-th highest (scaled) valuation, \ie, $\Delta_i = \tilde{q}(\sbb, a_{(1)}) - \tilde{q}(\sbb, a_{(i)})$ for $i=1,2,\ldots,m$. 
For ease of exposition, we will slightly abuse the notation and consider that there is a dummy action $a_{(m+1)}$ with valuation $\tilde{q}(\sbb, a_{(m+1)}) = -\infty$ and, consequently, $\Delta_{m+1} = +\infty$. Then, we can write
\begin{align*}
    \PP[\pi_{\Ccal}(\sbb, W; \eps) = \Ccal_{(i)}]
    &\stackrel{(*)}{=} \PP\left[ \left(\tilde{q}(\sbb, a_{(i)}) + W \geq \tilde{q}(\sbb, a_{(1)}) - \eps \right) \land \left(\tilde{q}(\sbb, a_{(i+1)}) + W < \tilde{q}(\sbb, a_{(1)}) - \eps \right) \right] \\
    &= \PP\left[\tilde{q}(\sbb, a_{(1)}) - \tilde{q}(\sbb, a_{(i)}) - \eps \leq W < \tilde{q}(\sbb, a_{(1)}) - \tilde{q}(\sbb, a_{(i+1)}) - \eps \right] \\
    &= \PP\left[\Delta_{i} - \eps \leq W < \Delta_{i+1} - \eps \right] \\
    &= \int_{\Delta_{i} - \eps}^{\Delta_{i+1} - \eps} g_W(w) dw
\end{align*}
where $(*)$ follows by the definition of the decision support policy in Eq.~\ref{eq:decision-support-policy} and $g_W$ is the probability density function of $W$. Now, without loss of generality, assume that $\eps < \eps'$. Then, it follows that
\begin{multline*}
    \left|  \PP[\pi_{\Ccal}(\sbb, W; \eps) = \Ccal_{(i)} ] -  \PP[\pi_{\Ccal}(\sbb, W; \eps') = \Ccal_{(i)} ] \right|
    = \left| \int_{\Delta_{i} - \eps}^{\Delta_{i+1} - \eps} g_W(w) dw - \int_{\Delta_{i} - \eps'}^{\Delta_{i+1} - \eps'} g_W(w) dw \right| \\
    \stackrel{(*)}{=} \left| \int_{\Delta_{i}}^{\Delta_{i+1}} g_W(x - \eps) dx - \int_{\Delta_{i}}^{\Delta_{i+1}} g_W(x - \eps') dx \right| \\
    \stackrel{(**)}{\leq} \int_{\Delta_{i}}^{\Delta_{i+1}} | g_W(x - \eps) - g_W(x - \eps') | dx,
\end{multline*}
where in $(*)$ we made the change of variable $x = w + \eps$ ($\eps'$) in the first (second) integral and in $(**)$ we used the triangle inequality.
Summing over all $i=1,2,\ldots,m$, we get that
\begin{multline}\label{eq:to-bound}
    \sum_{i=1}^{m} \left|  \PP[\pi_{\Ccal}(\sbb, W; \eps) = \Ccal_{(i)} ] -  \PP[\pi_{\Ccal}(\sbb, W; \eps') = \Ccal_{(i)} ] \right| 
    \leq \sum_{i=1}^{m} \int_{\Delta_{i}}^{\Delta_{i+1}} | g_W(x - \eps) - g_W(x - \eps') | dx \\
    = \int_{0}^{+\infty} | g_W(x - \eps) - g_W(x - \eps') | dx
    \stackrel{(*)}{=} \int_{-\eps'}^{+\infty} | g_W(y + (\eps' - \eps)) - g_W(y) | dy\\
    = \int_{-\eps'}^{0} | g_W(y + (\eps' - \eps)) - g_W(y) | dy + \int_{0}^{+\infty} | g_W(y + (\eps' - \eps)) - g_W(y) | dy \\
    \stackrel{(**)}{\leq} \int_{-\eps}^{\eps'-\eps} | g_W(y') | dy' + 
    \int_{-\eps'}^0 |g_W(y)| dy
    + \int_{0}^{+\infty} | g_W(y + (\eps' - \eps)) - g_W(y) | dy \\
    \stackrel{(***)}{=} \int_{0}^{\eps'-\eps} | g_W(y') | dy' + \int_{0}^{+\infty} | g_W(y + (\eps' - \eps)) - g_W(y) | dy,
\end{multline}
where in $(*)$ we made the change of variable $y = x - \eps'$, in $(**)$ we used the triangle inequality and made the change of variable $y' = y + (\eps' - \eps)$, and in $(***)$ we used the fact that $g_W(y) = 0$ for all $y < 0$ since $W$ follows a half-normal distribution.
Since the probability density function of a half-normal distribution is given by $g_W(y) = \frac{\sqrt{2}}{\sigma \sqrt{\pi}} \cdot e^{-y^2 / 2\sigma^2}$ for $y \geq 0$, it is easy to verify that $g_W(y)$ is upper bounded by $\frac{\sqrt{2}}{\sigma \sqrt{\pi}}$ and, therefore, the first term on the right-hand side of the above inequality is upper bounded by $\frac{\sqrt{2}}{\sigma \sqrt{\pi}} \cdot (\eps' - \eps)$.

To bound the second term, we use the fundamental theorem of calculus, which gives that, for any $y \geq 0$,
\begin{equation*}
    \left|g_W(y + (\eps' - \eps)) - g_W(y)\right| = \left|\int_{y}^{y + (\eps' - \eps)} g'_W(x) dx\right| \stackrel{(*)}{\leq} \int_{0}^{\eps' - \eps} \left|g'_W(\theta + y)\right|d\theta,
\end{equation*}
where in $(*)$ we made the change of variable $\theta = x - y$ and used the triangle inequality. Hence, we get
\begin{multline*}
    \int_{0}^{+\infty} | g_W(y + (\eps' - \eps)) - g_W(y) | dy 
    \leq \int_{0}^{+\infty} \int_{0}^{\eps' - \eps} |g'_W(\theta + y)| d\theta dy 
    = \int_{0}^{\eps' - \eps} \int_{0}^{+\infty} |g'_W(\theta + y)| dy d\theta \\
    \stackrel{(*)}{\leq} \int_{0}^{\eps' - \eps} \int_{0}^{+\infty} |g'_W(y)| dy d\theta
    = (\eps' - \eps) \cdot \int_{0}^{+\infty} |g'_W(y)| dy
    = (\eps' - \eps) \cdot \int_{0}^{+\infty} \frac{\sqrt{2}}{\sigma \sqrt{\pi}} \frac{y}{\sigma^2} \cdot e^{-y^2 / 2\sigma^2} dy \\
    \stackrel{(**)}{=} (\eps' - \eps) \cdot \frac{\sqrt{2}}{\sigma \sqrt{\pi}} \cdot \int_{0}^{+\infty}  e^{-u} du
    = (\eps' - \eps) \cdot \frac{\sqrt{2}}{\sigma \sqrt{\pi}},
\end{multline*}
where $(*)$ follows from the fact that $\theta, y, |g'_W| \geq 0$ and in $(**)$ we made the change of variable $u = y^2 / 2\sigma^2$. Having bounded both terms on the right-hand side of Eq.~\ref{eq:to-bound} with $\frac{\sqrt{2}}{\sigma \sqrt{\pi}} \cdot (\eps' - \eps)$, we can conclude that the proposition holds with $L_c = \frac{2\sqrt{2}}{\sigma \sqrt{\pi}}$.

\subsection{Proof of Proposition~\ref{prop:lipischitz-v}}

To prove the proposition, it suffices to show that, for any $\zb\in\Zcal$ and $\eps,\eps'\in[0,1]$, it holds that
\begin{equation}\label{eq:conditional}
    \left| v(\zb ; \eps) - v(\zb ; \eps') \right| \leq L\cdot \left|\eps-\eps'\right|.
\end{equation}
Once that is established, using Jensen's inequality, we are able to get
\begin{multline*}
   \left | \EE_{\Zb_{1}}[v(\Zb_{1};\eps)] - \EE_{\Zb_{1}}[v(\Zb_{1};\eps')] \right|
    \leq \EE_{\Zb_1}[ |v(\Zb_1;\eps)-v(\Zb_1;\eps')|] 
    = \int_{\Zcal} |v(\zb;\eps)-v(\zb;\eps')| \cdot f_{Z_1}(\zb) d\zb \\
    \leq L\cdot|\eps - \eps'| \int_{\Zcal} f_{Z_1}(\zb) d\zb 
    = L\cdot|\eps - \eps'|,
\end{multline*}
where $f_{\Zb_1}$ is the probability density function of the initial state $\Zb_1$.

The main idea of the proof is to first show a variant of Eq.~\ref{eq:conditional} for a finite horizon $T$, and then extend it to the infinite-horizon case by taking the limit $T \to \infty$.
For the finite-horizon case, our goal is to show that, for any horizon $T\geq 1$, there exists a constant $L_T > 0$ such that for all $\zb\in\Zcal$ and $\eps, \eps' \in [0,1]$ it holds that
\begin{equation}\label{eq:finite-horizon-lipschitz}
    |v_T(\zb;\eps) - v_T(\zb;\eps') | \leq L_T \cdot | \eps - \eps' |,
\end{equation}
where $v_T(\zb;\eps) = \EE_{\Wb, \Bb, \Ub} \left[ \sum_{t=0}^{T-1} \gamma^t \cdot R_{t+1} \given \Zb_{1} = \zb\right]$.

We will prove that Eq.~\ref{eq:finite-horizon-lipschitz} holds for any finite horizon $T$ by induction. We start with the base case $T=1$, where the quantity $v_1(\zb;\eps)$ corresponds to the immediate reward obtained by the human, depending on the state $\zb$ and its representation $\sbb = f_{S}(\zb)$, the action set $\Ccal$ provided by the decision support system, and the action $A$ they choose from that action set. Similarly to the proof of Proposition~\ref{prop:half-normal}, we will use the fact that the decision support policy $\pi_{\Ccal}$ can assign non-zero probability only to action sets $\Ccal$ that are of the form $\Ccal_{(i)} = \{ a_{(1)}, a_{(2)}, \ldots, a_{(i)} \}$, where $a_{(1)}, a_{(2)}, \ldots, a_{(m)}$ is the ranking of the $m$ actions based on the valuations $q(\sbb, a)$ of the AI agent. Then, we can write
\begin{equation*}
    v_1(\zb;\eps) = \sum_{i=1}^{m} \PP[\pi_{\Ccal}(\sbb, W; \eps) = \Ccal_{(i)} ] \sum_{a\in\Acal} \PP[A = a \given \Zb = \zb, \Ccal = \Ccal_{(i)}] \cdot f_R(\zb, a, \varnothing),
\end{equation*} 
where we have used $f_R(\zb, a, \varnothing)$ to denote the reward obtained by the human when they take an action $a$ in a state $\zb$ right before the end of the episode. Following from that, we get
\begin{align*}
    | v_1(\zb;\eps) -  v_1(\zb;\eps')|
    &\leq \sum_{i=1}^{m} \left |\PP[\pi_{\Ccal}(\sbb, W; \eps) = \Ccal_{(i)} ]  - \PP[\pi_{\Ccal}(\sbb, W; \eps') = \Ccal_{(i)} ] \right | \\
    &\qquad\cdot\sum_{a\in\Acal} \PP[A = a \given \Zb = \zb, \Ccal = \Ccal_{(i)}] \cdot |f_R(\zb, a, \varnothing)|\\
    &\stackrel{(*)}{\leq} \sum_{i=1}^{m} \left |\PP[\pi_{\Ccal}(\sbb, W; \eps) = \Ccal_{(i)} ]  - \PP[\pi_{\Ccal}(\sbb, W; \eps') = \Ccal_{(i)} ] \right |  \cdot  r_{max}\\
    &\stackrel{(**)}{\leq} r_{max} \cdot L_c \cdot | \eps - \eps'|,
\end{align*}
where $(*)$ follows from the fact that the rewards are bounded by $r_{max}$, and $(**)$ follows from Proposition~\ref{prop:half-normal}. Therefore, the base case holds with $L_1 = r_{max} \cdot L_c$.

Now, we will prove the inductive step. Assume that Eq.~\ref{eq:finite-horizon-lipschitz} holds for $T=n-1$ for some natural number $n\geq 2$, that is, there exists $L_{n-1} > 0$ such that, for all $\zb \in \Zcal$ and $\eps, \eps' \in [0,1]$, it holds that
\begin{equation*}
    |v_{n-1}(\zb;\eps) - v_{n-1}(\zb;\eps') | \leq L_{n-1} \cdot | \eps - \eps' |.
\end{equation*}
We will show that it also holds for $T = n$. First, we can write
\begin{align*}
    v_n(\zb;\eps) 
    &= \sum_{i=1}^{m} \PP[\pi_\Ccal(\sbb, W;\eps) = \Ccal_{(i)} ]
    \sum_{a\in\Acal}  \PP[a\given \zb, \Ccal_{(i)}] \int_{\Zcal} \PP[\zb' \given \zb, a] \cdot \left( f_R(\zb,a,\zb')  + \gamma \cdot v_{n-1}(\zb';\eps) \right) d\zb' \\
    &= \underbrace{\sum_{i=1}^{m} \PP[\pi_\Ccal(\sbb, W;\eps) = \Ccal_{(i)} ]
    \sum_{a\in\Acal}  \PP[a\given \zb, \Ccal_{(i)}] \int_{\Zcal} \PP[\zb' \given \zb, a] f_R(\zb,a,\zb') d\zb'}_{\text{Immediate reward term } (\dagger \,;\, \eps)} \\
    &\qquad + \underbrace{\gamma \sum_{i=1}^{m} \PP[\pi_\Ccal(\sbb, W;\eps) = \Ccal_{(i)} ]
    \sum_{a\in\Acal}  \PP[a\given \zb, \Ccal_{(i)}] \int_{\Zcal} \PP[\zb' \given \zb, a] \cdot v_{n-1}(\zb';\eps) d\zb'}_{\text{Future discounted cumulative reward term } (\ddagger \,;\, \eps)},
\end{align*}
where, for brevity, we have denoted as $\PP[a\given \zb, \Ccal_{(i)}]$ the probability that the human takes action $a$ in state $\zb$ when they are provided with action set $\Ccal_{(i)}$ and as $\PP[\zb' \given \zb, a]$ the transition probability from state $\zb$ to state $\zb'$ when the human takes action $a$.
Further, for ease of exposition, we will bound separately the absolute differences of the two terms $(\dagger)$ and $(\ddagger)$ resulting from $\eps, \eps'$.
Starting with the immediate reward term, we have
\begin{align*}
    | (\dagger \,;\, \eps) - (\dagger \,;\, \eps') |
    &\stackrel{(*)}{\leq} \sum_{i=1}^{m} \left |\PP[\pi_\Ccal(\sbb, W;\eps) = \Ccal_{(i)} ] - \PP[\pi_\Ccal(\sbb, W;\eps') = \Ccal_{(i)} ] \right | \\
    &\qquad\cdot\sum_{a\in\Acal}  \PP[a\given \zb, \Ccal_{(i)}] \int_{\Zcal} \PP[\zb' \given \zb, a] \cdot | f_R(\zb,a,\zb') | d\zb'\\
    &\stackrel{(**)}{\leq} \sum_{i=1}^{m} \left |\PP[\pi_\Ccal(\sbb, W;\eps) = \Ccal_{(i)} ] - \PP[\pi_\Ccal(\sbb, W;\eps') = \Ccal_{(i)} ] \right | r_{max}
    \stackrel{(***)}{\leq} r_{max} \cdot L_c \cdot | \eps - \eps' |,
\end{align*}
where $(*)$ follows from the triangle inequality, $(**)$ follows from the fact that the rewards are bounded by $r_{max}$, and $(***)$ follows from Proposition~\ref{prop:half-normal}.

Next, we will bound the difference of the future discounted cumulative reward terms. We can write it as
\allowdisplaybreaks
\begin{align*}
    | (\ddagger \,;\, \textcolor{red}{\eps}) - (\ddagger \,;\, \textcolor{blue}{\eps'}) |
    &= \Bigg| \gamma \sum_{i=1}^{m} \PP[\pi_\Ccal(\sbb, W;\textcolor{red}{\eps}) = \Ccal_{(i)} ]
    \sum_{a\in\Acal}  \PP[a\given \zb, \Ccal_{(i)}] \int_{\Zcal} \PP[\zb' \given \zb, a] \cdot v_{n-1}(\zb';\textcolor{red}{\eps}) d\zb'\\
    &\qquad - \gamma \sum_{i=1}^{m} \PP[\pi_\Ccal(\sbb, W;\textcolor{blue}{\eps'}) = \Ccal_{(i)} ]
    \sum_{a\in\Acal}  \PP[a\given \zb, \Ccal_{(i)}] \int_{\Zcal} \PP[\zb' \given \zb, a] \cdot v_{n-1}(\zb';\textcolor{blue}{\eps'}) d\zb'\\
    &\qquad \pm \gamma \sum_{i=1}^{m} \PP[\pi_\Ccal(\sbb, W;\textcolor{blue}{\eps'}) = \Ccal_{(i)} ]
    \sum_{a\in\Acal}  \PP[a\given \zb, \Ccal_{(i)}] \int_{\Zcal} \PP[\zb' \given \zb, a] \cdot v_{n-1}(\zb';\textcolor{red}{\eps}) d\zb' \Bigg| \\
    &\stackrel{(*)}{\leq} \gamma \sum_{i=1}^{m} \left| \PP[\pi_\Ccal(\sbb, W;\textcolor{red}{\eps}) = \Ccal_{(i)} ] - \PP[\pi_\Ccal(\sbb, W;\textcolor{blue}{\eps'}) = \Ccal_{(i)} ] \right| \\
    &\qquad\qquad\cdot\sum_{a\in\Acal}  \PP[a\given \zb, \Ccal_{(i)}] \int_{\Zcal} \PP[\zb' \given \zb, a] \cdot | v_{n-1}(\zb';\textcolor{red}{\eps}) | d\zb' \\
    &\qquad + \gamma \sum_{i=1}^{m} \PP[\pi_\Ccal(\sbb, W;\textcolor{blue}{\eps'}) = \Ccal_{(i)} ] \\
    &\qquad\qquad\cdot\sum_{a\in\Acal}  \PP[a\given \zb, \Ccal_{(i)}] \int_{\Zcal} \PP[\zb' \given \zb, a] \cdot | v_{n-1}(\zb';\textcolor{red}{\eps}) - v_{n-1}(\zb';\textcolor{blue}{\eps'}) | d\zb' \\
    &\stackrel{(**)}{\leq} \gamma \cdot L_c \cdot | \textcolor{red}{\eps} - \textcolor{blue}{\eps'} | \cdot \frac{r_{max}}{1 - \gamma} + \gamma \cdot L_{n-1} \cdot | \textcolor{red}{\eps} - \textcolor{blue}{\eps'} |,
\end{align*}
where we have used red color for $\textcolor{red}{\eps}$ and blue color for $\textcolor{blue}{\eps'}$ for ease of exposition.
Here, $(*)$ follows from the triangle inequality and in $(**)$ we used Proposition~\ref{prop:half-normal}, the induction hypothesis, and the fact that
\begin{equation*}
    | v_{n-1}(\zb';\eps) | \leq \sum_{t=0}^{n-2} \gamma^t \cdot r_{max} \leq \sum_{t=0}^{\infty} \gamma^t \cdot r_{max} = \frac{r_{max}}{1 - \gamma}.
\end{equation*}

Combining the bounds for the two terms, we can conclude that
\begin{align*}
    | v_n(\zb;\eps) -  v_n(\zb;\eps')| 
    &\leq \left( r_{max} \cdot L_c + \gamma \cdot L_c \cdot \frac{r_{max}}{1 - \gamma} + \gamma \cdot L_{n-1} \right) \cdot | \eps - \eps' | \\
    &= \left( L_c \cdot \frac{r_{max}}{1 - \gamma} + \gamma \cdot L_{n-1} \right) \cdot | \eps - \eps' |,
\end{align*}
which yields the closed form expression
\begin{equation*}
    L_n = L_c \cdot r_{max} \cdot \left[\gamma^{n-1} + \frac{1 - \gamma^{n-1}}{1 - \gamma}\cdot \frac{1}{1-\gamma} \right] = L_c \cdot r_{max} \cdot \frac{\gamma^n(\gamma-2) +1}{(1-\gamma)^2}.
\end{equation*}

As a result, we have shown that Eq.~\ref{eq:finite-horizon-lipschitz} holds for any finite horizon $T$.
Now, we will extend the result to the infinite-horizon case.
First, it is easy to see that the limit $\lim_{T\to\infty} v_T(\zb;\eps)$ is finite for all $\zb \in \Zcal$ and $\eps \in [0,1]$ since $|v_T(\zb;\eps)| \leq \sum_{t=0}^{T-1} \gamma^t \cdot r_{max} \leq \frac{r_{max}}{1 - \gamma}$ for all $T \geq 1$.
Therefore, we can write
\begin{multline*}
| v(\zb;\eps) -  v(\zb;\eps') |
= \left| \lim_{T\to\infty} v_T(\zb;\eps) - \lim_{T\to\infty} v_T(\zb;\eps') \right|
= \left| \lim_{T\to\infty} \left( v_T(\zb;\eps) - v_T(\zb;\eps') \right) \right| \\
\leq \lim_{T\to\infty} \left| v_T(\zb;\eps) - v_T(\zb;\eps') \right|
\stackrel{(*)}{\leq} \lim_{n\to\infty} L_n \cdot | \eps - \eps' |,
\end{multline*}
where in $(*)$ we used the result for the finite horizon case. Finally, it is easy to verify that $\lim_{n\to\infty} L_n = \frac{L_c \cdot r_{max}}{(1-\gamma)^2}$, which concludes the proof of the proposition for the infinite horizon case with $L = \frac{L_c \cdot r_{max}}{(1-\gamma)^2}$.

\subsection{Proof of Proposition~\ref{prop:simple-regret}}\label{app:simple-regret}

For ease of exposition, here we assume that all observed payoffs take values in $[0,1]$. Note that this is without loss of generality, since the reward at each time step lies in the bounded interval $[-r_{max}, r_{max}]$ and, consequently, the (discounted) cumulative reward is also bounded.
Therefore, this assumption has an effect only on constants in the simple regret guarantees of Algorithm~\ref{algo:best-arm-identification} that we derive next.

We will prove the proposition by distinguishing between two types of events: a ``clean'' event that holds with high probability and the complementary ``bad'' event, similarly to the analysis of other bandit algorithms~\citep{slivkins2019introduction}.
Our goal is to first establish a lower bound on the probability of the clean event, and then show that, if the clean event holds, then the algorithm returns an arm $\eps_{ALG}$ with a sufficiently small simple regret, \ie, $p(\eps^*) - p(\eps_{ALG})$.

Formally, let $E$ denote the clean event where, for all iterations of the algorithm and all active intervals, the estimates of the average payoffs for all arms are sufficiently accurate, \ie,
\begin{equation*}
    E = \bigcap_{k=1}^{k_{max}} \bigcap_{I \in \Ical_k} \left\{ |\hat{p}(\eps_I) - p(\eps_I)| \leq l_k \right\},
\end{equation*}
where $k_{max}$ is the maximum number of iterations of the algorithm, $l_k = 2^{-k}$ is the length of the intervals at iteration $k$, and $\eps_I$ is the midpoint of interval $I$. 

We will start by showing that the number of iterations $k_{max}$ is bounded.
Note that, at each iteration $k$, the algorithm samples $n_k = 2^{k\beta}$ times at each active interval, while the total number of samples available to the algorithm is $n$.
Therefore, in the worst case where the algorithm eliminates all but one interval at each iteration, the maximum number of iterations $k_{max}$ has to satisfy
\begin{equation*}
    n \geq \sum_{k=1}^{k_{max}} n_k = \sum_{k=1}^{k_{max}} 2^{k\beta} = \frac{2^{\beta}(2^{k_{max}\beta}-1)}{2^{\beta}-1} \Rightarrow k_{max} := \left\lceil \frac{\log_2(n(2^{\beta}-1)+1)}{\beta} \right\rceil .
\end{equation*}

Now, we will establish a lower bound on the probability of the clean event $E$.
Since for each interval $I \in \Ical_k$, the estimate $\hat{p}(\eps_I)$ is obtained by averaging $n_k$ payoff samples, Hoeffding's inequality implies that
\begin{equation*}
    \PP[|\hat{p}(\eps_I) - p(\eps_I)| > l_k] \leq 2\exp(-2n_k l_k^2).
\end{equation*}
Moreover, since the samples drawn to estimate the payoffs at the midpoints of the intervals in each iteration are independent of the algorithm's choices in previous iterations, the events $\{ |\hat{p}(\eps_I) - p(\eps_I)| \leq l_k \}$ for all $I \in \Ical_k$ are independent. Therefore, the probability of the clean event $E$ can be bounded as follows:
\begin{multline*}
    \PP[E]
    = \PP\left[\bigcap_{k=1}^{k_{max}} \bigcap_{I \in \Ical_k} \{ |\hat{p}(\eps_I) - p(\eps_I)| \leq l_k \}\right] 
    = 1 - \PP\left[\bigcup_{k=1}^{k_{max}} \bigcup_{I \in \Ical_k} \{ |\hat{p}(\eps_I) - p(\eps_I)| > l_k \}\right] \\
    \stackrel{(*)}{\geq} 1 - \sum_{k=1}^{k_{max}} \sum_{I \in \Ical_k} \PP[|\hat{p}(\eps_I) - p(\eps_I)| > l_k] 
    \geq 1 - \sum_{k=1}^{k_{max}} \sum_{I \in \Ical_k} 2\exp(-2n_k l_k^2)\\
    = 1 - 2\sum_{k=1}^{k_{max}} |\Ical_k| \exp(-2^{k(\beta-2) - 1})
    \stackrel{(**)}{\geq} 1 - 2\sum_{k=1}^{k_{max}} 2^{k} \exp(-2^{k(\beta-2) - 1}),
\end{multline*}
where $(*)$ follows from the union bound and $(**)$ follows from the fact that $[0,1]$ can be covered by at most $2^k$ intervals of length $l_k = 2^{-k}$.

Next, we will show that, if the clean event $E$ holds, then the algorithm returns an arm with a sufficiently small simple regret.
We start by showing that the optimal arm $\eps^*$ is never eliminated during the execution of the algorithm.
For the sake of contradiction, assume that, at some iteration $k$, the interval $I^*_k$ containing the optimal arm $\eps^*$ is eliminated.
By the definition of the elimination rule, this would mean that there exists an interval $I \in \Ical_k$ such that
\begin{equation*}
    \hat{p}(\eps_{I}) - \hat{p}(\eps_{I^*_k}) > (2 + L/2)l_k.
\end{equation*}
However, since we are considering the case where the clean event $E$ holds, we have that
\begin{align*}
    \hat{p}(\eps_{I}) - \hat{p}(\eps_{I^*_k})
    &= \hat{p}(\eps_{I}) - p(\eps_{I}) + p(\eps_{I}) - p(\eps^*) + p(\eps^*) - p(\eps_{I^*_k}) + p(\eps_{I^*_k}) - \hat{p}(\eps_{I^*_k}) \\
    &\leq p(\eps_{I}) - p(\eps^*) + |\hat{p}(\eps_{I}) - p(\eps_{I})| + |p(\eps^*) - p(\eps_{I^*_k})| + |\hat{p}(\eps_{I^*_k}) - p(\eps_{I^*_k})| \\
    &\stackrel{(*)}{\leq} p(\eps_{I}) - p(\eps^*) + l_k + L \cdot \frac{l_k}{2} + l_k \\
    &\stackrel{(**)}{\leq} (2+L/2) \cdot l_k,
\end{align*}
where $(*)$ follows from the definition of the clean event and the Lipschitz continuity of $p$, and $(**)$ follows from the fact that $\eps^*$ is optimal.
This leads to a contradiction and, hence, intervals containing the optimal arm $\eps^*$ are never eliminated during the execution of the algorithm, as long as the clean event $E$ holds.

We will now establish an upper bound on the gap between the payoff of the optimal arm $\eps^*$ and the payoffs of the arms in the non-eliminated intervals at each iteration $k$.
Consider a (non-eliminated) interval $I \in \Ical_k$ and an arm $\eps \in I$.
Since $I$ was not eliminated at iteration $k$, it holds that
\begin{equation*}
    \hat{p}(\eps_{I^*_k}) - \hat{p}(\eps_I) \leq (2+ L/2)l_k
\end{equation*}
and, since the clean event $E$ holds, we have
\begin{equation*}
    \hat{p}(\eps_{I^*_k}) \geq p(\eps_{I^*_k}) - l_k,\,\,\,\text{and}\,\,\, \hat{p}(\eps_I) \leq p(\eps_I) + l_k.
\end{equation*}
Combining the above inequalities, we obtain
\begin{equation*}
    p(\eps_{I^*_k}) - p(\eps_I) \leq (2+ L/2)l_k + 2l_k = (4 + L/2)l_k,
\end{equation*}
and, using the Lipschitz continuity of $p$, for any arm $\eps \in I$, we have
\begin{align*}\label{eq:optimality-gap-2}
    p(\eps^*) - p(\eps)
    &\leq p(\eps^*) - p(\eps_{I^*_k}) + p(\eps_{I^*_k}) - p(\eps_I) + p(\eps_I) - p(\eps) \\    
    &\leq L\cdot \frac{l_k}{2} + (4 + L/2)l_k + L\cdot \frac{l_k}{2} \\
    &= (4 + 3L/2)l_k. \numberthis
\end{align*}

The above inequality implies that the union of all intervals $I \in \Ical_k$ is a subset of the set of arms with a payoff within $(4 + 3L/2)l_k$ of the optimal payoff.
We will now use the definition of the zooming dimension to bound the number of active intervals $|\Ical_k|$ at iteration $k$.
Consider a radius $\rho = (4 + 3L/2)l_k$. Then, we know that there exist constants $\lambda$ and $d$ such that, the set of arms with a payoff within $\rho$ of the optimal payoff can be covered by $N = \lambda \cdot \rho^{-d}$ sets of diameter at most $\rho$.
Since the intervals in $\Ical_k$ are disjoint and have length $l_k = \frac{\rho}{4 + 3L/2}$, it has to hold that
\begin{equation*}
    |\Ical_k| \cdot l_k \leq N \cdot \rho 
    \Rightarrow |\Ical_k| \leq (4 + 3L/2) \cdot N
    \Rightarrow |\Ical_k| \leq (4 + 3L/2)^{(1-d)} \cdot \lambda \cdot l_k^{-d}.
\end{equation*}

Let $K$ denote the last iteration of the algorithm, that is, the iteration after which the algorithm exhausted its budget $n$.
Since the algorithm terminated right after iteration $K$, it has to hold that
\begin{equation*}
    n < \sum_{k=1}^{K+1} |\Ical_k| n_k \leq \sum_{k=1}^{K+1} (4 + 3L/2)^{(1-d)} \cdot \lambda \cdot l_k^{-d} \cdot n_k = (4 + 3L/2)^{(1-d)} \cdot \lambda \cdot \sum_{k=1}^{K+1} 2^{k(d+\beta)}.
\end{equation*}

Then, we can bound the above sum as follows:
\begin{align*}
    \sum_{k=1}^{K+1} 2^{k(d+\beta)} 
    = 2^{d+\beta} \cdot \frac{2^{(K+1)(d+\beta)} - 1}{2^{d+\beta} - 1}
    \leq 2^{d+\beta} \cdot \frac{2^{(K+1)(d+\beta)}}{2^{d+\beta} - 1}
    = 2^{d+\beta} \cdot \frac{2^{d+\beta}\cdot l_K^{-(d+\beta)}}{2^{d+\beta} - 1},
\end{align*}
which, after rearranging the terms, yields
\begin{equation*}
    l_K \leq n^{\frac{-1}{d+\beta}} \cdot \left( (4 + 3L/2)^{(1-d)} \cdot \lambda \cdot \frac{4^{d+\beta}}{2^{d+\beta} - 1} \right)^{\frac{1}{d+\beta}}.
\end{equation*}
Combining the above inequality with Eq.~\ref{eq:optimality-gap-2} and observing that the arm $\eps_{ALG}$ returned by our algorithm comes from an interval not eliminated at iteration $K$, we get that
\begin{equation}\label{eq:full-regret}
    p(\eps^*) - p(\eps_{ALG}) \leq n^{\frac{-1}{d+\beta}} \cdot \left(4 + 3L/2\right)^{\frac{1+\beta}{d+\beta}} \cdot \left( \frac{\lambda \cdot 4^{d+\beta}}{2^{d+\beta} - 1} \right)^{\frac{1}{d+\beta}} = \Ocal \left(n^{\frac{-1}{d+\beta}}\right).
\end{equation}
This concludes the proof of the proposition with
\begin{align*}
    \delta_n = 2\sum_{k=1}^{k_{max}} 2^{k} \exp(-2^{k(\beta-2) - 1})
    \,\,\,\text{and}\,\,\,
    k_{max} = \left\lceil \frac{\log_2(n(2^{\beta}-1)+1)}{\beta} \right\rceil.
\end{align*}

\clearpage
\newpage

\section{Additional experimental details}
\label{app:experiment-details}

\begin{figure}
    \centering
    \includegraphics[width=0.7\linewidth]{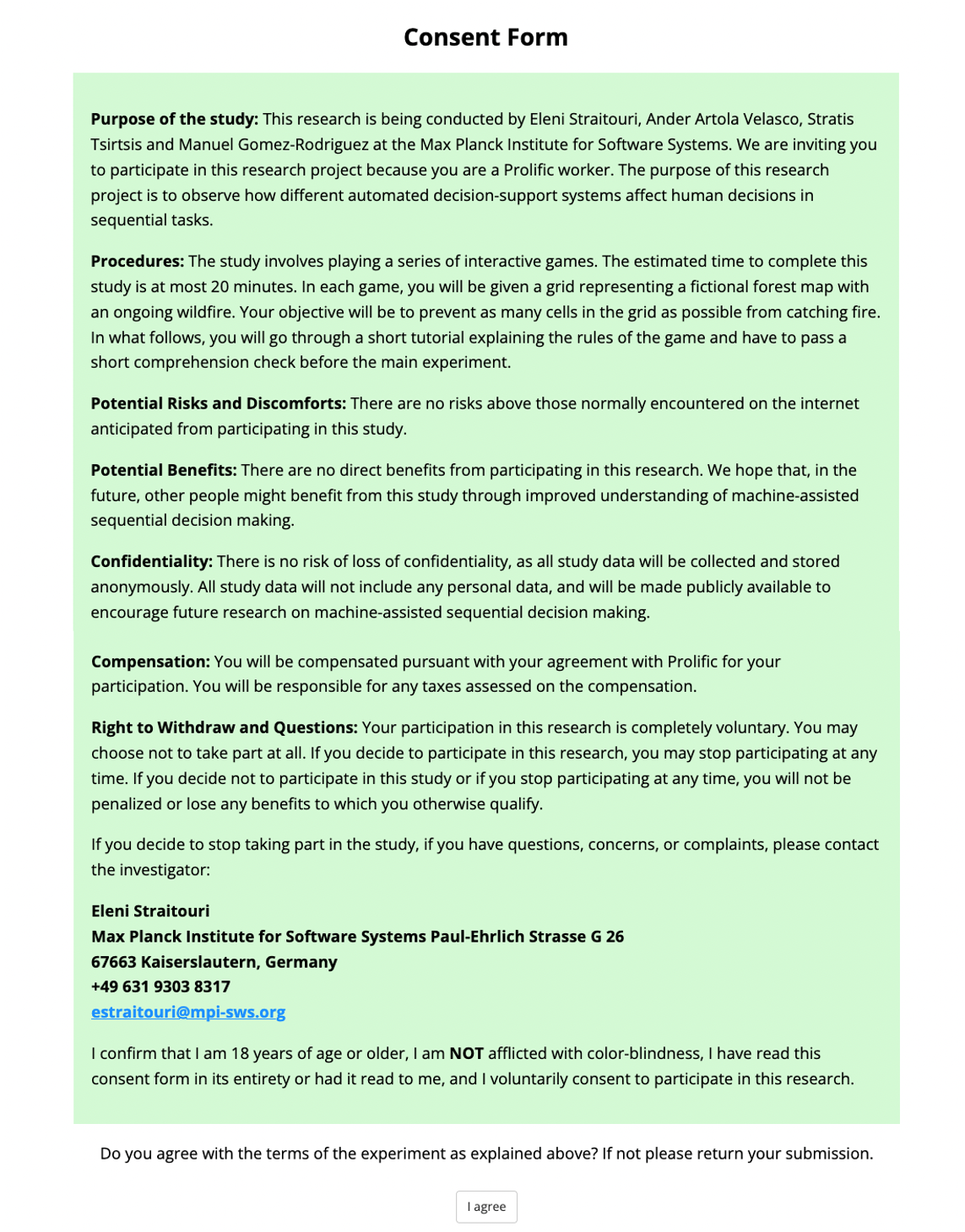}
    \caption{The consent form that Prolific workers had to read and agree to in order to participate in our study.}
    \label{fig:consent}
\end{figure}

\begin{figure}
    \centering
    \includegraphics[width=\linewidth]{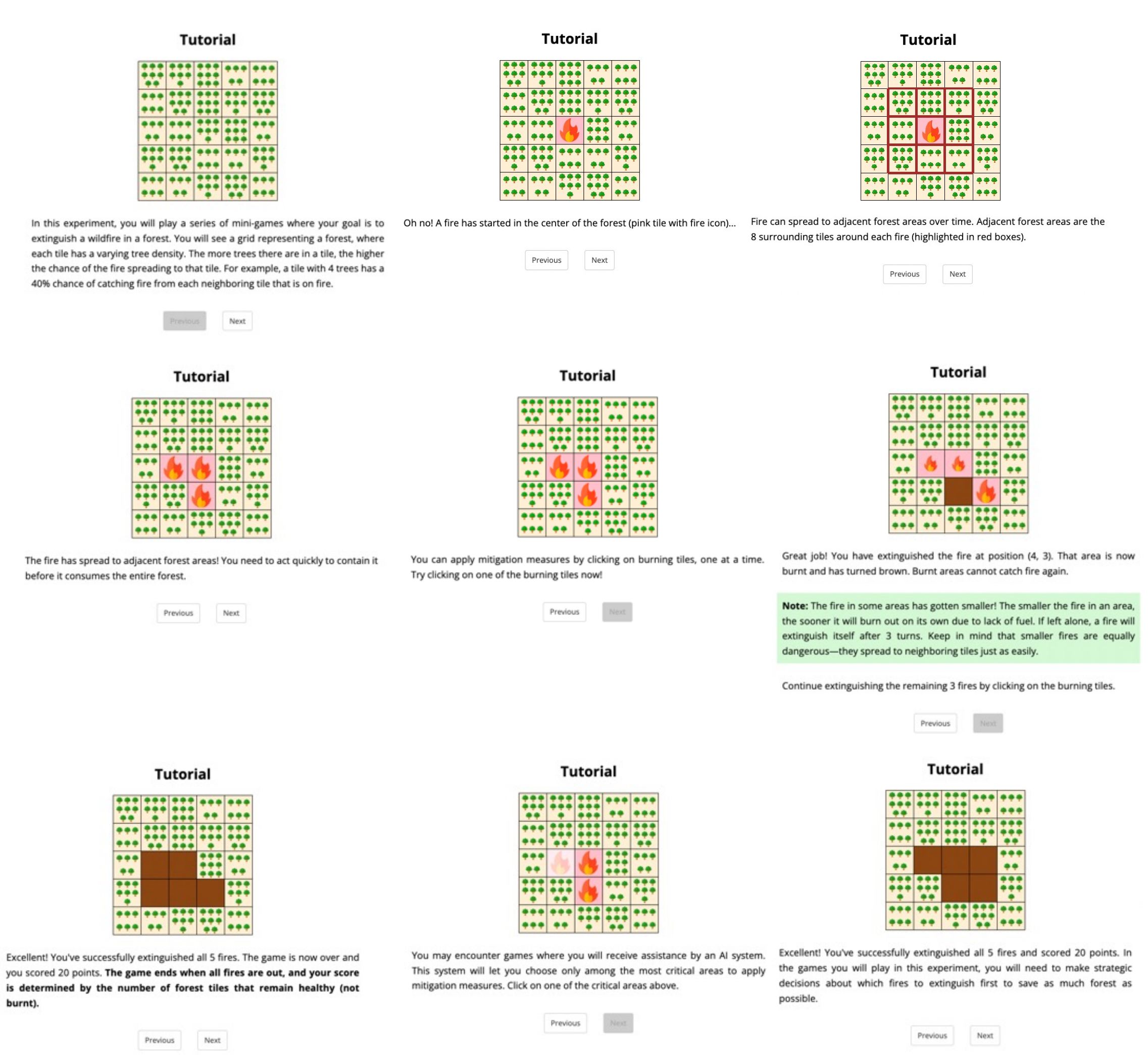}
    \caption{Snapshots of the tutorial that the human subject study participants had to go through to understand and familiarize with the game procedures. The snapshots are presented in their order of appearance in the tutorial from left to right and top to bottom.}
    \label{fig:tutorial}
\end{figure}
\begin{figure}
    \centering
    \includegraphics[width=0.6\linewidth]{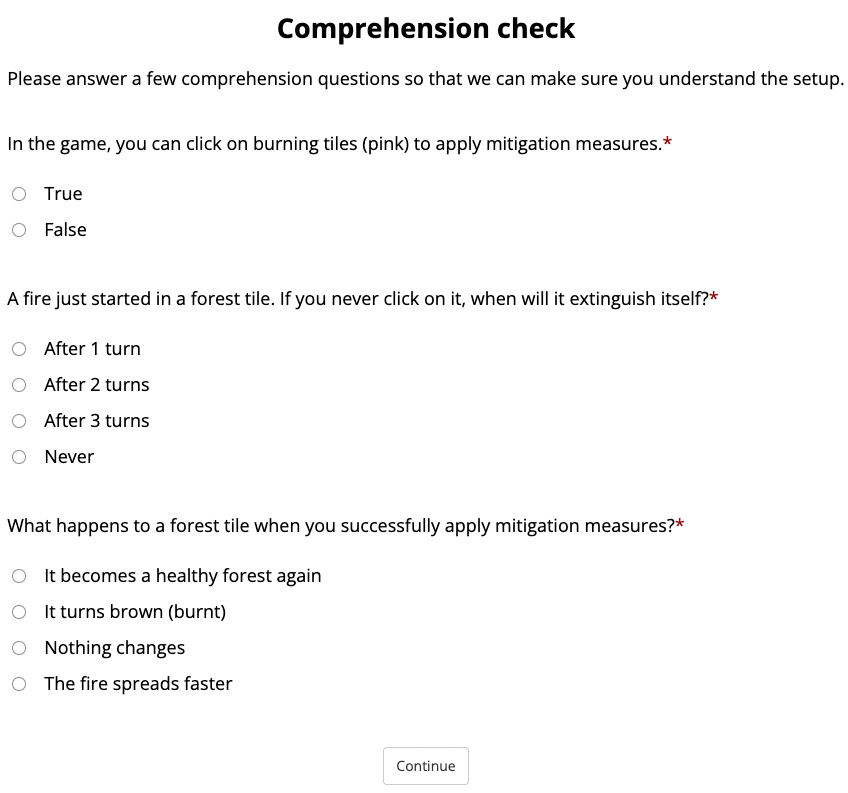}
    \caption{The comprehension test that participants had to pass after the tutorial to proceed with our study.}
    \label{fig:comprehension-test}
\end{figure}
\xhdr{Further details on human subject study} Each participant in our human subject study provided their consent by filling the form shown in Figure~\ref{fig:consent}. After filling this form, each participant went through the tutorial shown in Figure~\ref{fig:tutorial}, followed by a comprehension check shown in Figure~\ref{fig:comprehension-test}.

\xhdr{Fire spread dynamics} To allow for stochastic fire spread dynamics, we assign a density parameter $p_{i,j}$ to each grid tile with coordinates $(i,j)$, which controls how easy it is for the tile to catch fire and remains constant throughout the game.
Each of these parameters takes a value $p_{i,j}\in\{0.1, 0.2, \ldots, 0.9\}$, which we generate using a Gaussian random field, and we visualize using $1, 2, \ldots, 9$ trees, respectively, in the corresponding tile (see Fig.~\ref{fig:game-example}).

At each time step $t$, we decide whether to set the status of each \texttt{healthy} tile $(i,j)$ to \texttt{burning} proportionally to its density $p_{i,j}$ and the number $|\Ncal_{i,j,t}|$, where $\Ncal_{i,j,t}$ is the set of all \texttt{burning} tiles adjacent to $(i,j)$ at time $t$.
Specifically, we consider that each tile $(i',j')\in\Ncal_{i,j,t}$ makes an attempt to spread the fire to the tile $(i,j)$ with probability $p_{i,j}$, and the tile $(i,j)$ catches fire if at least one of these attempts is successful.
Formally, we sample a random variable $X_{i,j,t} \sim \text{Binomial}(|\Ncal_{i,j,t}|, p_{i,j})$ and set the status of tile $(i,j)$ to \texttt{burning} if $X_{i,j,t} \geq 1$.

\begin{figure}
    \centering
    \includegraphics[width=0.6\textwidth]{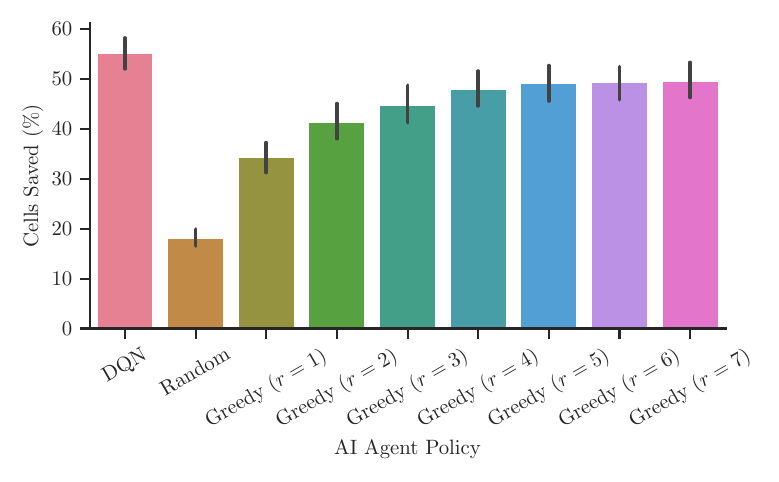}
    \caption{Comparative performance of the trained DQN agent against AI agents implementing heuristic policies.}
    \label{fig:baselines}
\end{figure}

\xhdr{Deep Q-Network}
We build our AI Agent using reinforcement learning and, during training, we optimize it to operate autonomously in our wildfire mitigation environment without any human in the loop.
Specifically, we train a Deep Q-Network (DQN)~\citep{mnih2015human} with a double Q-learning approach~\citep{van2016deep} to provide valuations $q(\sbb,a)$ such that they approximate the expected (discounted) cumulative reward the AI agent can obtain by taking action $a$ while seeing a representation $\sbb = f_S(\zb)$ and following the optimal policy $\pi_M^*$ thereafter.
To align the AI agent's objective with that of the participants of the human subject study (\ie, saving as many forest tiles as possible), we set the AI agent's (training) reward at each time step $t$ and discount factor $\gamma$ in a similar manner as in human subject study---we set the reward equal to the negative number of tiles that turned from \texttt{healthy} to \texttt{burning} in each time step and $\gamma=0.99$. In addition, we provide a large negative reward equal to the negative total number of \texttt{burnt} tiles at the end of each game to speed up training.

To better enable the AI agent to leverage the spatial structure of the environment, we use a network architecture based solely on convolutional layers---four convolutional layers with kernel sizes of $3 \times 3$, $3 \times 3$, $5 \times 5$, and $7 \times 7$, respectively, and $32$, $32$, $64$, and $64$ filters, respectively, followed by ReLU activations.
In addition, to speed up training, we use action masking to ensure that the AI agent only applies water mitigation measures to tiles on the firefront, that is, tiles that are \texttt{burning} and have at least one \texttt{healthy} neighboring tile.

For training the DQN, we generate $2{,}000$ random game instances and perform $75$ simulations for each instance, resulting in a total of $150{,}000$ training games.
To ensure that the game instances used for training are sufficiently diverse in terms of their difficulty, we first evaluate the performance achieved by AI agents implementing several heuristic policies (see below) on a large set of randomly generated game instances by measuring the percentage of forest tiles that they manage to maintain \texttt{healthy} at the end of the game.
Then, we categorize the game instances into five difficulty levels based on the performance range (\eg, $20\% - 40\%$) of the best heuristic policy (Greedy with radius $r=7$, see below), and we create a balanced training set by keeping $400$ game instances from each difficulty level.
We repeat the same procedure to create a balanced test set of $200$ game instances, which we use to evaluate the performance of our AI agent using the trained DQN against AI agents using heuristic policies.
Figure~\ref{fig:baselines} shows their comparative performance on the test set, and a detailed description of the heuristic policies can be found below:
\begin{itemize}
    \item \textbf{Random policy}: The agent randomly selects a tile that is on the firefront and applies a water mitigation measure to it.
    \item \textbf{Greedy policy with radius $r\in\{1, 2, \ldots, 7\}$}: For each tile $(i,j)$ on the firefront, the policy computes the product of densities on each path of length $r$ starting from $(i,j)$ and ending in an \texttt{healthy} tile $(i',j')$ and adds them up. The agent then applies water mitigation measures to the tile $(i,j)$ that maximizes this sum. In the special case of $r=1$, the agent simply selects the tile on the firefront whose \texttt{healthy} neighbors have the highest total density.
\end{itemize}

\xhdr{Decision support policies} We select the $\sigma$ and $\eps$ values we use in our human subject study based on simulations we conducted prior to the human subject study. 
Specifically, we 
evaluated our decision support policies for $\eps$ values from $0.0$ to $1.0$ with step $0.01$ and $\sigma \in \{0.001, 0.01, 0.05, 0.1\}$ using simulated humans, which we elaborate on in further detail in the next paragraph.
Based on the simulation results, we identified that our simulated humans under the assistance of our decision support policies achieved the highest average cumulative reward for $\eps\in[0, 0.3]$ and $\sigma=0.01$.
In our human subject study presented in Section~\ref{sec:experiments}, we deploy our decision support policies for each $\eps$ value in the aforementioned range with step $0.01$ and for $10$ more values evenly spread outside this range (\ie, in $[0.3, 1.0]$) for completeness.

%
%

To build our simulated humans, we conducted a  small-scale pilot study prior to our large-scale human subject study.
Therein, we asked $30$ volunteers to play $200$ instances of our wildfire mitigation game on their own (\ie, without any AI assistance).
For each set of instances starting from the same state, we measured the average (discounted) cumulative reward achieved by the volunteers and that achieved by AI agents implementing a set of heuristic policies including the  greedy policies as described above, as well as softmax versions of the respective greedy policy.\footnote{A softmax heuristic policy with radius $r$ computes the same sum as the respective greedy policy for each tile $(i,j)$ on the firefront and applies a softmax function over those sums. The agent then selects which tile they will apply water mitigation measures on by drawing a random sample from the categorical distribution induced by the values of this softmax function.}
Then, we build our simulated humans such that, for each set of instances starting from the same state, they play the game on their own by implementing the heuristic policy that most closely matched the respective average (discounted) cumulative reward achieved by the volunteers in that set of instances.
To set the human policy $\pi_H$ that our simulated humans follow in the case where they play games under the assistance of our decision support policies, we consider that they implement the same heuristic policy in the respective set of instances, with the only difference that the respective random sampling or argmax operation that leads to the action selection is taken over the (restricted) set of tiles in the action set given by the decision support policy at each time step, rather than over the entire firefront.

\xhdr{Licenses and implementation details} We will publicly release our code and human subject study data under the Creative Commons Attribution License 4.0 (CC BY 4.0). To implement our algorithm and execute our experiments, we use \texttt{Python} 3.11.2 along with the open-source libraries \texttt{PyTorch} 2.8.0 (Modified BSD License), \texttt{NumPy} 2.3.2 (BSD License), \texttt{Pandas} 2.3.2 (BSD License). For our human subject study we develop a web application, where we implement the fronend with the \texttt{jsPsych} framework (MIT License), and the backend with the open-source libraries \texttt{FastAPI} 0.115.12 (MIT License) and \texttt{Uvicorn} 0.34.3 (BSD License).

\end{document}